\documentclass[nonacm, sigplan]{acmart}

\usepackage{hyperref}   
\usepackage{hyperxmp}   
\usepackage{booktabs}
\usepackage{subcaption} 
\usepackage{enumitem}
\usepackage{algorithm}
\usepackage[noend]{algpseudocode}
\usepackage{makecell}
\usepackage{multicol}
\usepackage{multirow}
\usepackage[table]{xcolor}

\usepackage{setspace}
\usepackage{adjustbox}

\usepackage{etoolbox}
\usepackage{tcolorbox}

\newcommand{\THISWORK}{{\fontfamily{lmss}\selectfont
Sphinx}}

\begin{document}


\title[Sphinx]{\THISWORK: Efficiently Serving Novel View Synthesis using Regression-Guided Selective Refinement}

\author{Yuchen Xia}
\affiliation{%
  \institution{University of Illinois Urbana-Champaign}
  \city{Urbana}
  \state{IL}
  \country{USA}
}
\email{yxia28@illinois.edu}

\author{Souvik Kundu}
\affiliation{%
  \institution{Intel}
  \city{Los Angeles}
  \state{CA}
  \country{USA}
}
\email{souvikk.kundu@intel.com}

\author{Mosharaf Chowdhury}
\affiliation{%
  \institution{University of Michigan}
  \city{Ann Arbor}
  \state{MI}
  \country{USA}
}
\email{mosharaf@umich.edu}

\author{Nishil Talati}
\affiliation{%
  \institution{University of Illinois Urbana-Champaign}
  \city{Urbana}
  \state{IL}
  \country{USA}
}
\email{nishil@illinois.edu}

\begin{abstract}
Novel View Synthesis (NVS) generates novel views of a scene from a sparse set of input images and is a key component in applications such as immersive rendering and 3D content generation.
Existing NVS approaches broadly exhibit an efficiency--quality tradeoff: regression-based models generate novel views with relatively low latency but can struggle with perceptual quality, particularly in challenging regions or regions not visible in the input views (disoccluded regions), while diffusion-based models produce higher-quality and temporally consistent views at substantially greater computational cost.
This motivates a hybrid approach that leverages the complementary strengths of regression and diffusion to bridge this efficiency--quality gap.

We present \textbf{\THISWORK}, a \textbf{\textit{training-free}} hybrid inference and serving system for NVS that combines regression-level efficiency with diffusion-level quality.
\THISWORK\ first uses fast regression-based NVS to generate coarse novel views, which provide a strong initialization for diffusion and avoid expensive generation from pure noise. It then selectively applies diffusion refinement where additional computation is needed, reducing the denoising workload while preserving the quality benefits of diffusion-based NVS.
To efficiently serve the resulting heterogeneous requests, \THISWORK\ further introduces an SLO-aware serving framework that dynamically allocates GPU resources based on each request's remaining computation and urgency.
Evaluated across diverse NVS models, datasets, and serving configurations, \THISWORK\ achieves an average \textbf{2.3}$\times$ inference speedup (up to \textbf{2.7}$\times$) with less than \textbf{10\%} perceptual degradation, while its SLO-aware scheduler achieves up to \textbf{10}$\times$ higher SLO attainment than FCFS scheduling across hardware platforms.

\end{abstract}


\maketitle 

\section{Introduction} \label{section:introduction}

\begin{figure}[t]
    \centering
    \includegraphics[width=\linewidth]{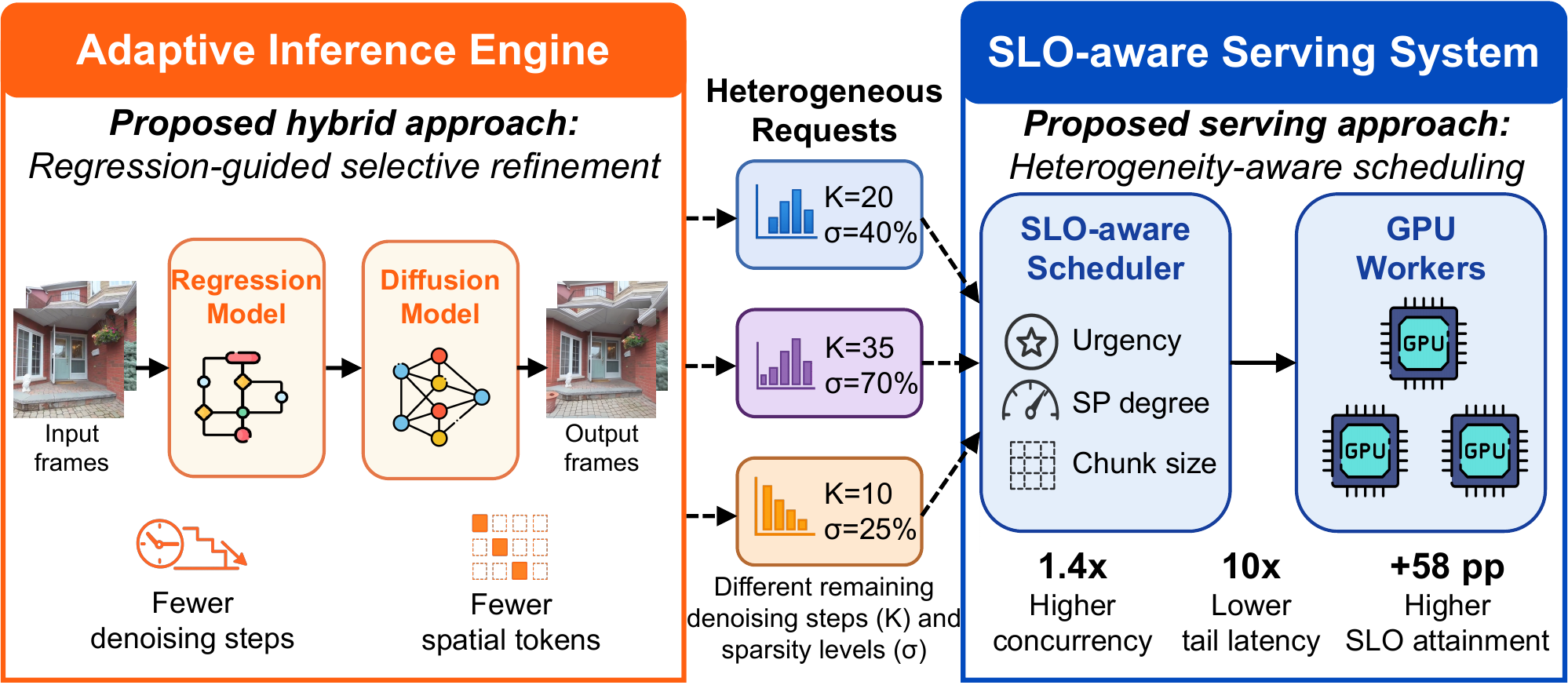}
    \vspace{-0.8cm}
    \caption{\textit{Overview of \THISWORK. Adaptive inference reduces denoising and spatial computation, while SLO-aware serving efficiently schedules the resulting heterogeneous requests.}}
    \vspace{-2mm}
    \label{fig:overview}
\end{figure}

Novel view synthesis (NVS) generates unseen viewpoints from a limited set of input images and is becoming a key primitive for emerging 3D applications, including AR/VR, digital twins, real-time telepresence, robotics, and autonomous-driving simulation~\cite{zhang2025advancesfeedforward3dreconstruction,telealoha,ma2025novelview,tian2025viewinvariant}. 
These applications increasingly demand interactive or repeated view generation, making inference latency and serving efficiency critical. Existing NVS methods broadly fall into \textit{regression-based} and \textit{diffusion-based} approaches~\cite{berian2025modern}. Regression-based models~\cite{mvsplat,depthsplat,liu2025mvsgaussian,pixelsplat} predict target views in a single feed-forward pass with low latency, but often produce oversmoothing or artifacts in uncertain and disoccluded regions. Diffusion-based models~\cite{seva,Wu2026framecrafter,4DiM,gen3c,motionctrl,viewcrafter} instead use strong generative priors and iterative denoising to produce high-quality, temporally consistent views at substantially higher computational cost. \textit{These model families therefore expose a fundamental trade-off between inference efficiency and visual quality.}

Their complementary strengths motivate a hybrid approach: use a fast regression model to initialize target views, then apply diffusion only where needed. However, fixed amount of denoising and uniform spatial refinement ignore variations in regression quality, leaving this opportunity underexploited. Regression quality varies across scenes and frames, with errors often concentrated in uncertain regions, so outputs require different amounts of diffusion computation. Uniform diffusion refinement either wastes computation on reliable content or under-refines difficult regions. Efficient hybrid inference must therefore adapt \emph{how much} diffusion computation is performed and \emph{where}, while preserving perceptual quality and temporal coherence. Prior distillation-based hybrid approaches~\cite{wang2025videoscenedistillingvideodiffusion} require additional training and may need retraining for new models or datasets.

This adaptivity creates a system-level challenge. Requests may differ in remaining denoising depth and the fraction of spatial tokens requiring diffusion refinement, producing heterogeneous costs even under the same model and resolution. Additionally, Sequence Parallelism (SP) is widely used to accelerate compute-bound diffusion inference across multiple GPUs ~\cite{fang2024xdit, SwiftFusion}, but introduces a latency--concurrency trade-off: more GPUs reduce one request's latency while limiting capacity for others. Under online workloads, request-oblivious scheduling can let expensive requests block lighter or more urgent ones, while static GPU allocations cannot adapt as requests progress. Efficient serving must therefore decide \emph{which} requests to execute, \emph{how many} GPUs to allocate, and \emph{how much} denoising work to perform before rescheduling. This motivates our central question: \textit{Can we jointly adapt inference computation and GPU allocation to approach regression-level efficiency and diffusion-level quality while maintaining high SLO attainment?}

We present \THISWORK, a \textbf{training-free algorithm--system co-design} for efficient NVS inference and serving. As illustrated in Figure~\ref{fig:overview}, \THISWORK\ warm-starts diffusion from regression outputs. It then adapts refinement across scenes, frames, timesteps, and spatial regions: quality-aware noise scheduling determines denoising depth, latent reuse preserves temporal consistency under partial denoising, spatially selective execution restricts computation to uncertain regions, and scene-aware control adapts refinement to input characteristics. These decisions expose each request's remaining denoising depth and selected-token ratio, summarizing its remaining computation. We propose using these signals as an algorithm--system interface to the serving layer. Our SLO-aware scheduler combines them with offline latency profiles and, at denoising-chunk boundaries, dynamically updates request priority, SP allocation, chunk size (the number of denoising steps per scheduling interval), and GPU placement based on remaining work and SLO urgency.

We evaluate \THISWORK\ on three NVS datasets---RE10K~\cite{re10k}, DL3DV~\cite{dl3dv}, and ACID~\cite{acid}---using two representative diffusion-based NVS models, SEVA~\cite{seva} and FrameCrafter~\cite{Wu2026framecrafter}.We measure quality with MUSIQ~\cite{musiq}, LPIPS~\cite{lpips}, PSNR, and SSIM~\cite{ssim}, alongside inference latency, tail latency, and SLO attainment. Across both models and three datasets, \THISWORK\ achieves an average \textbf{2.3}$\times$ speedup (up to \textbf{2.7}$\times$) over full diffusion inference with less than \textbf{10\%} perceptual degradation, while its SLO-aware scheduler improves SLO attainment by up to \textbf{58\%} over request-oblivious FCFS serving across A40 and H200 GPUs. Our key contributions are:
\begin{itemize}[leftmargin=*]
    \item A training-free adaptive regression--diffusion NVS pipeline that warm-starts diffusion and adapts refinement across scenes, frames, timesteps, and spatial regions.
    
    \item An algorithm--system interface that exposes request-specific denoising depth and spatial sparsity as online cost signals for heterogeneity-aware serving.

    \item An SLO-aware serving system that uses these signals to jointly adapt request priority, SP allocation, chunk size, and GPU placement according to remaining computation and SLO urgency, achieving up to \textbf{10}$\times$ higher SLO attainment than request-oblivious FCFS serving.
\end{itemize}

\vspace{-0.23cm}

\section{Background and Motivation} \label{section:Background}

\subsection{Novel View Synthesis (NVS)} \label{subsec:nvs_background}

\begin{figure}[t]
    \centering
    \includegraphics[width=\linewidth]{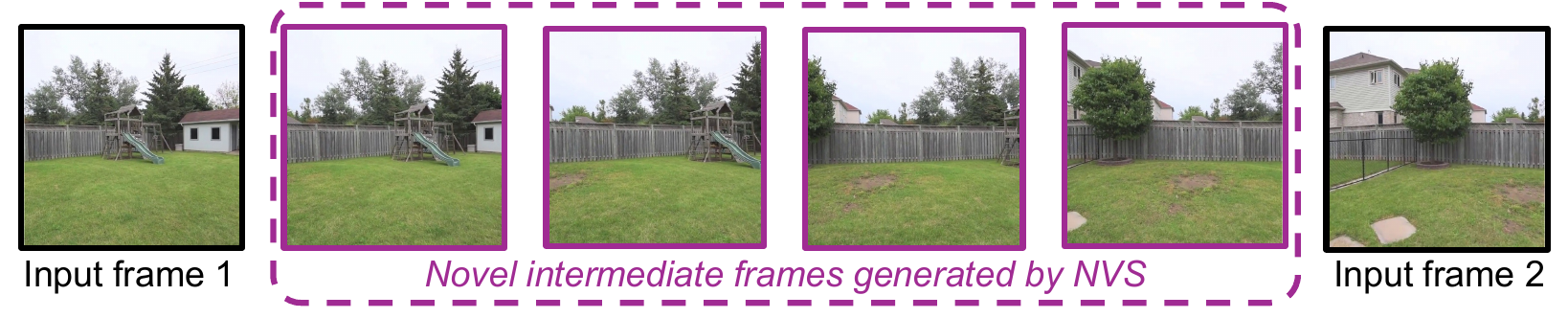}
    \vspace{-0.9cm}
    \caption{\textit{Example NVS task: two input camera frames are used to generate four intermediate novel views between them.}}
    \label{fig:nvs_background}
    \vspace{-0.3cm}
\end{figure}

NVS aims to generate unseen views of a scene from a limited set of input images with known camera parameters~\cite{liu2025survey}. The goal is to produce geometrically consistent and photorealistic renderings from novel viewpoints, or along a smooth camera trajectory to form a temporally coherent video sequence. Figure~\ref{fig:nvs_background} illustrates an example in which two input views are used to synthesize intermediate frames. NVS is a core problem in vision and graphics, with applications in 3D reconstruction, AR/VR, and immersive media creation.

The primary algorithmic challenge is jointly modeling scene geometry and appearance from limited observations. Traditional methods based on explicit geometry and image-based rendering often fail under occlusions or sparse inputs, while modern neural approaches learn implicit or generative representations end-to-end to produce photorealistic and view-consistent outputs.

\subsection{Regression- and Diffusion-Based NVS} \label{subsec:nvs_algo_model}

Existing NVS methods fall into two categories: regression-based and diffusion-based approaches~\cite{berian2025modern}.

\textbf{Regression-based models} deterministically map input views and camera poses to target views. NeRF-based methods~\cite{nerf} optimize per-scene 3D representations with high geometric accuracy but costly optimization, while feed-forward methods such as MVSplat~\cite{mvsplat}, PixelSplat~\cite{pixelsplat}, MVSGaussian~\cite{liu2025mvsgaussian} and DepthSplat~\cite{depthsplat} enable real-time rendering but degrade under sparse inputs or large viewpoint changes.

\textbf{Diffusion-based models} learn a conditional generative distribution over target views using strong diffusion priors. Methods such as MotionCtrl~\cite{motionctrl}, 4DiM~\cite{4DiM}, GEN3C~\cite{gen3c}, ViewCrafter~\cite{viewcrafter}, FrameCrafter~\cite{Wu2026framecrafter}, and SEVA~\cite{seva} produce high-quality, temporally consistent results even from sparse inputs, but incur substantial computational cost because they repeatedly execute the denoising network.

\subsection{Analysis of the Performance--Quality Trade-off}
\label{subsec:perf_quality_tradeoff}

\begin{figure}[t]
    \centering
    \includegraphics[width=0.8\linewidth]{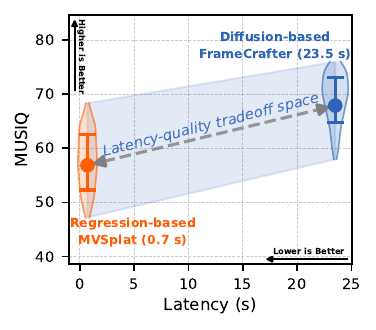}
    \vspace{-0.7cm}
    \caption{Latency--quality gap between regression- and diffusion-based NVS, measured on a NVIDIA H200 GPU.}
    \label{fig:tradeoff}
    \vspace{-0.55cm}
\end{figure}

Regression-based NVS methods emphasize geometric reconstruction and computational efficiency, whereas diffusion-based approaches prioritize perceptual fidelity. Recent regression-based models~\cite{mvsplat, zhang2025transplat, wang2024freesplat} extract image features, lift them into 3D representations such as Gaussians, and render target views in a feed-forward manner, enabling fast inference, while diffusion-based models synthesize target views through iterative denoising, where repeated network passes progressively refine the latent representation~\cite{you2025nvs, watson2024controlling}. Although this iterative process improves visual quality, it incurs substantially higher computation and latency. Figure~\ref{fig:tradeoff} illustrates this fundamental \textbf{\textit{performance--quality trade-off}} using MVSplat~\cite{mvsplat} and FrameCrafter~\cite{Wu2026framecrafter} as representative regression- and diffusion-based models, respectively, over 400 requests from DL3DV. The violin plots show the distribution of per-request MUSIQ scores~\cite{musiq}, while the markers indicate the average quality. MVSplat completes inference in only \textbf{0.7\,s} with an average MUSIQ of approximately \textbf{57}, whereas FrameCrafter improves the average score to approximately \textbf{69} but requires \textbf{23.5\,s}, resulting in a \textbf{33.6$\times$} increase in latency. This large gap between efficient regression and high-quality diffusion motivates a natural question: \textit{Can we retain the efficiency of regression-based NVS while leveraging the generative strength of diffusion for high-quality outputs?}

\section{Adaptive Inference Engine}
\label{sec:inference_design}

\begin{figure}[t]
    \centering
    \includegraphics[width=\linewidth]{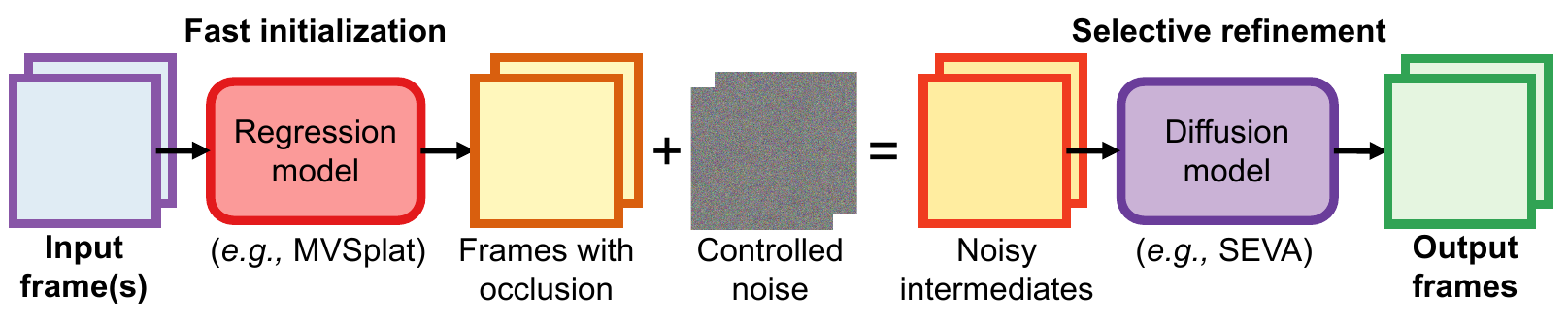}
    \vspace{-0.9cm}
    \caption{Hybrid inference combining regression-based fast initialization with diffusion-based selective denoising.}
    \vspace{-0.3cm}
    \label{fig:proposed_hybrid_approach}
\end{figure}




To answer this question, we propose a hybrid inference approach that uses a fast regression model to generate initial target views and apply diffusion to refine them toward higher perceptual quality. However, applying a fixed amount of denoising to every output overlooks substantial variation in regression quality across scenes, target frames, and spatial regions. Different outputs—and even different regions within the same output—therefore require different amounts of diffusion refinement. We thus adapt diffusion computation to the quality and uncertainty of each regression prediction.

We propose \THISWORK, an adaptive hybrid inference approach that combines regression-level efficiency with diffusion-level quality by determining both \emph{how much} and \emph{where} diffusion refinement should be performed. As illustrated in Figure~\ref{fig:proposed_hybrid_approach}, \THISWORK\ first uses a regression model to generate coarse target views with per-pixel confidence information, providing a structured initialization for diffusion. \THISWORK\ then injects controlled noise into the regression output and starts denoising from an intermediate timestep instead of pure noise, allowing it to skip early denoising steps and reduce diffusion computation. It further restricts diffusion refinement only to low-confidence frames and spatial regions while reusing reliable content. However, realizing this adaptivity is nontrivial. The regression output provides no direct indication of how much diffusion is needed to recover the quality of full denoising, especially without access to the ground-truth target view. Moreover, diffusion models jointly process information across frames, so allowing different frames to begin denoising at different timesteps can disrupt temporal interactions between frames at different stages of refinement. Spatial adaptivity introduces a different systems challenge: prediction errors are irregularly distributed, and skipping reliable regions only reduces latency if the resulting sparsity can be efficiently mapped to GPU execution. Finally, the reliability of regression predictions varies across scenes, making a fixed refinement policy difficult to apply consistently.

These challenges give rise to four design goals. \textbf{DG1: Adaptive quality--computation balance.} \THISWORK\ must determine the appropriate denoising depth without access to ground-truth target views. \textbf{DG2: Temporal consistency.} It must preserve temporal coherence when different frames undergo different amounts of denoising. \textbf{DG3: Spatially selective refinement.} It must translate spatial uncertainty into GPU-efficient sparse execution. \textbf{DG4: Scene-aware quality--performance control.} It must account for scene-dependent variation in regression quality when selecting the refinement policy. We next describe how \THISWORK\ addresses each of these design goals.

\subsection{Quality-Aware Noise Scheduling}
\label{sec:noise_scheduling}

\begin{figure}[t]
    \centering
    \begin{subfigure}[t]{0.7\linewidth}
        \centering
        \includegraphics[width=\linewidth]{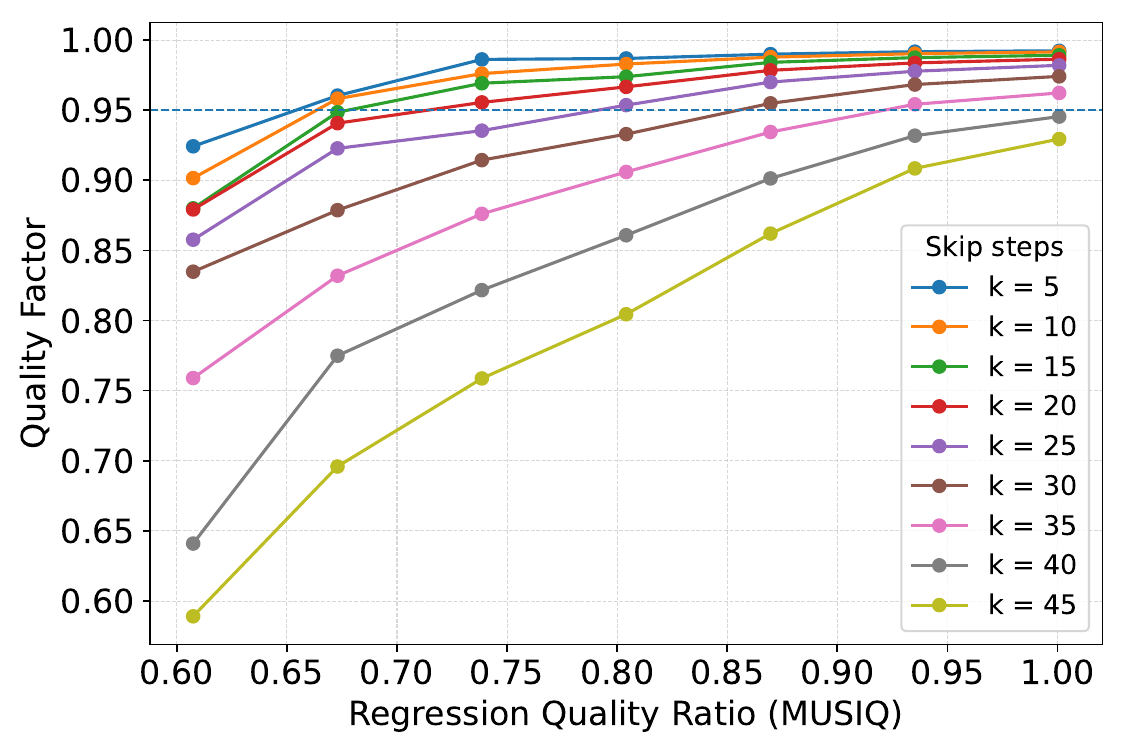}
        \caption{}
        \label{fig:musiq_curves}
    \end{subfigure}
    \hfill
    \begin{subfigure}[t]{0.29\linewidth}
        \centering
        \includegraphics[width=\linewidth]{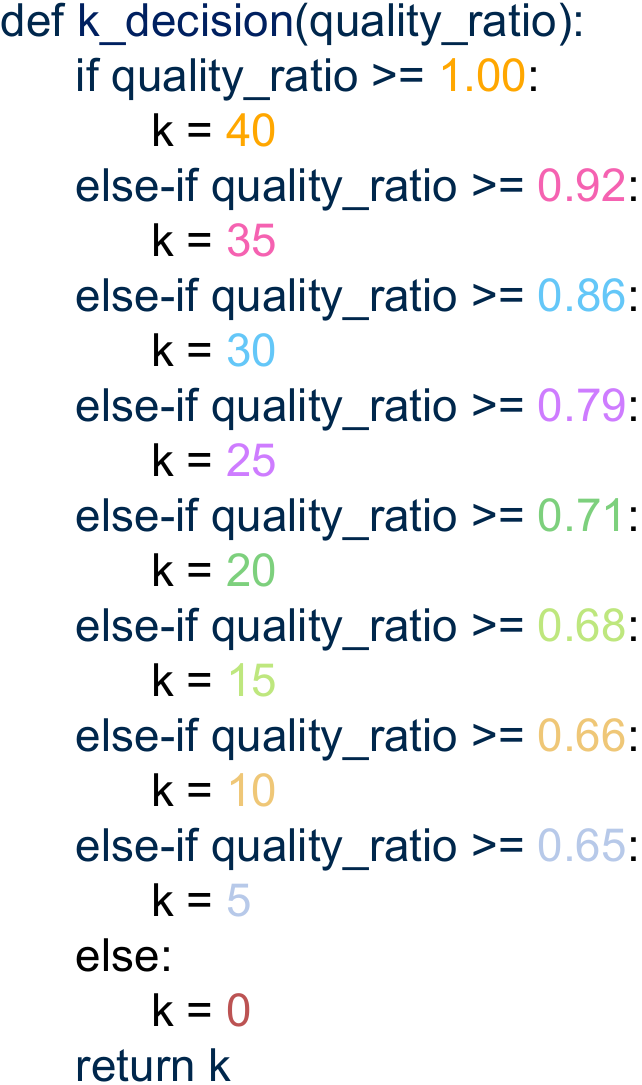}
        \caption{}
        \label{fig:k_logic}
    \end{subfigure}
    \vspace{-0.5cm}
    
    \caption{(a) Relationship between regression-quality ratio and resulting quality factor across different $k$; (b) decision logic for adaptive $k$ selection.}
    \vspace{-0.4cm}
    \label{fig:combined_fig}
\end{figure}

To address \textbf{DG1: adaptively balance quality and computation}, \THISWORK\ determines how much diffusion refinement each regression output requires. Starting diffusion earlier provides more denoising computation to correct regression errors, whereas starting later reduces latency but leaves less opportunity for refinement. We therefore employ quality-aware noise scheduling to select the denoising start point according to an explicit \textit{quality target}.

Let $Q_{\text{full}}$ denote the perceptual quality of an image produced by full diffusion, and let $Q_{\text{\THISWORK}}(k)$ represent the quality of our output when refinement begins at denoising step $k$. We define
{
\setlength{\abovedisplayskip}{3pt}
\setlength{\belowdisplayskip}{3pt}
\setlength{\abovedisplayshortskip}{3pt}
\setlength{\belowdisplayshortskip}{3pt}
\begin{equation}
Q_{\text{\THISWORK}}(k) \ge \alpha \cdot Q_{\text{full}},
\qquad 0 < \alpha \le 1,
\label{eq:quality-constraint}
\end{equation}
}
which requires each generated frame to achieve at least a fraction $\alpha$ of full-diffusion quality. For example, $\alpha=0.95$ targets at least 95\% of the perceptual quality of an image generated entirely by diffusion.

Selecting $k$ requires an image-quality estimate during inference. Full-reference metrics such as SSIM and PSNR are unsuitable because ground-truth target images are unavailable. We therefore use the non-reference metric MUSIQ~\cite{musiq}, which predicts perceptual quality directly from image content.

For each regression output, we obtain a quality score $\widehat{Q}_{\text{reg}}$. Absolute MUSIQ values vary with content and lighting, so \THISWORK\ instead evaluates the ratio between $\widehat{Q}_{\text{reg}}$ and an estimated reference quality. We approximate the reference by computing MUSIQ scores on the input views and interpolating the expected quality at each target viewpoint:
{
\setlength{\abovedisplayskip}{3pt}
\setlength{\belowdisplayskip}{3pt}
\begin{equation}
\begin{aligned}
Q^\star &= c_0 + (c_1 - c_0) \cdot f(t), \\[4pt]
f(t) &=
\begin{cases}
t^{\gamma}, & c_1 \ge c_0,\\[4pt]
1 - (1 - t)^{\gamma}, & c_1 < c_0,
\end{cases}
\end{aligned}
\label{eq:power-interp}
\end{equation}
}
where $c_0$ and $c_1$ are the MUSIQ scores of the two input frames, $t\in[0,1]$ is the normalized target-frame position, and $\gamma\in(0,1]$ controls interpolation sensitivity. A larger $\gamma$ produces a stricter estimated reference quality for intermediate frames.

We empirically determine the appropriate starting step $k$ under the constraint in Eq.~\eqref{eq:quality-constraint}. For a set of scenes, we generate regression and diffusion outputs and perform partial refinement from $k\in\{5,10,15,20,25,30\}$ within a 50-step diffusion process. Figure~\ref{fig:musiq_curves} plots the relationship between the regression-quality ratio and the final quality relative to full diffusion. Each curve corresponds to a different $k$ and is obtained from 2,000 RE10K~\cite{re10k} scenes using SEVA~\cite{seva} as the diffusion model and MVSplat~\cite{mvsplat} as the regression model. Figure~\ref{fig:k_logic} shows the resulting decision logic for $\alpha=0.95$. We cap $k$ at 40, ensuring the final 10 steps run densely to preserve consistency and avoid quality degradation.

Because regression quality varies across target viewpoints, this policy assigns different denoising depths across frames. This avoids redundant work but leaves only a subset of frames active at some timesteps. We next describe how \THISWORK\ preserves cross-frame consistency during partial execution.

\subsection{Temporally Consistent Partial Denoising}
\label{sec:temporal_consistency}

\begin{figure}[t]
    \centering
    \includegraphics[width=\linewidth]{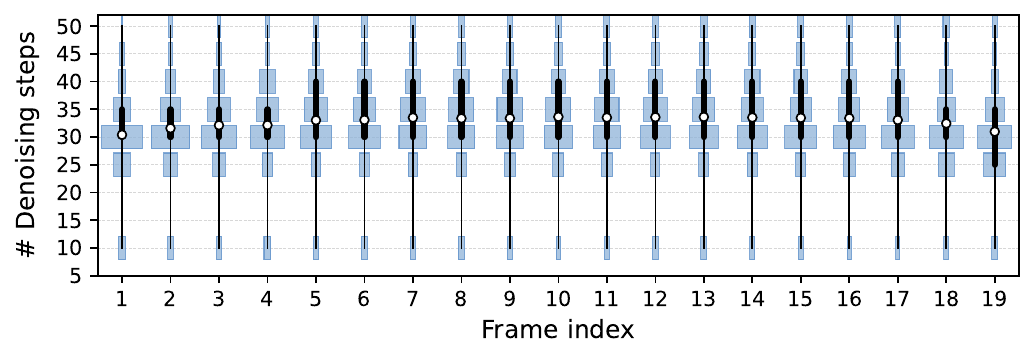}
    \vspace{-1cm}
    \caption{Distribution of selected denoising steps across target frames on the RE10K dataset; bar width indicates frequency, black lines denote the 25th--75th percentile range, and white dots mark the mean.}
    \label{fig:avg_k}
    \vspace{-0.5cm}
\end{figure}

To address \textbf{DG2: maintain temporal coherence under partial denoising}, \THISWORK\ preserves cross-frame consistency even when different frames execute different amounts of diffusion refinement. As shown in Figure~\ref{fig:avg_k}, denoising depth varies across frames and scenes, with frames farther from the input views typically requiring more denoising steps. Consequently, different subsets of frames remain active at different timesteps. Naively excluding inactive frames from computation can disrupt temporal attention and introduce flickering or structural inconsistency.

\begin{figure}[!t]
    \centering
    \includegraphics[width=0.8\linewidth]{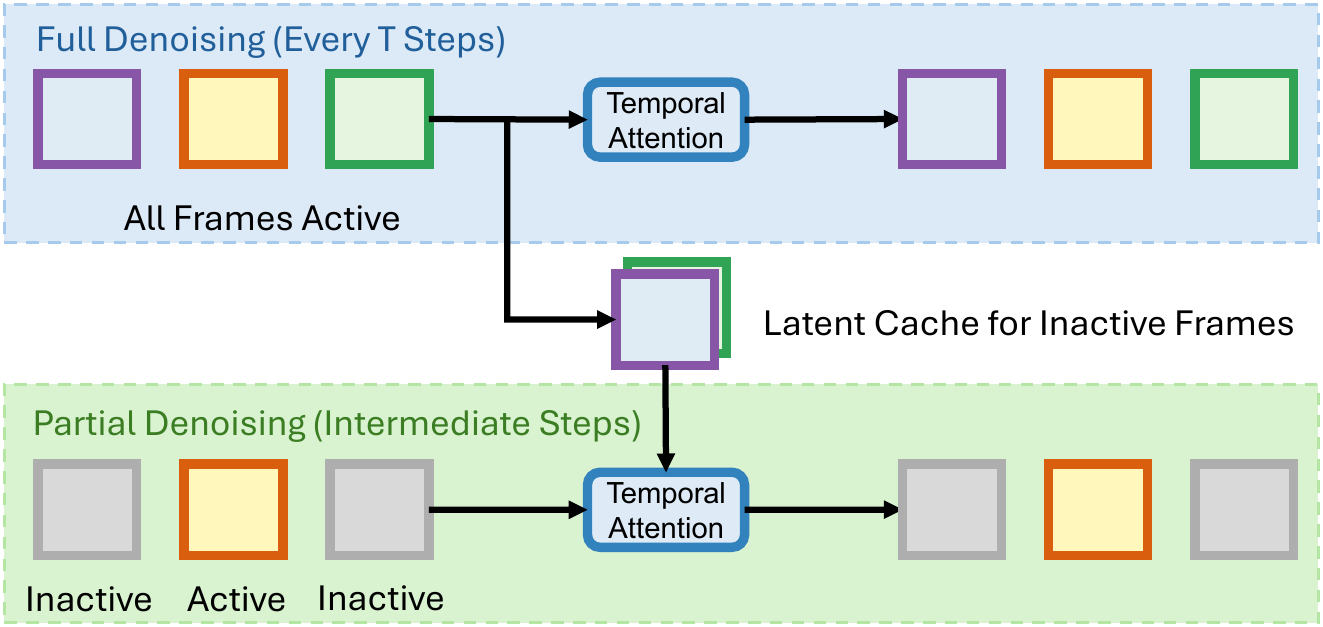}
    \vspace{-0.4cm}
    \caption{Temporal-cache design. We cache temporal-attention latents at a full denoising step $X_t$ and reuse them during the next $(T\!-\!1)$ partial steps for inactive frames.}
    \vspace{-0.2cm}
    \label{fig:cache}
\end{figure}

In diffusion models, convolutional and spatial-attention layers operate independently per frame and can be applied only to active frames. Temporal attention, however, aggregates information across all frames and requires complete temporal context, making partial-frame denoising difficult. To address this challenge, we propose a temporal latent cache that enables partial denoising while preserving temporal context (Figure~\ref{fig:cache}). Every $T$ steps, a full denoising pass refreshes the cached temporal-attention latents. During the following $(T-1)$ partial steps, inactive frames reuse these cached latents rather than recomputing them, reducing redundant computation while maintaining temporal consistency.

Temporal partial denoising reduces computation across frames and timesteps. Within active frames, regression errors are also localized, creating a second opportunity to restrict diffusion to regions requiring correction.

\subsection{Spatially Selective Refinement}

\begin{figure}[t]
\centering
\includegraphics[width=0.6\linewidth]{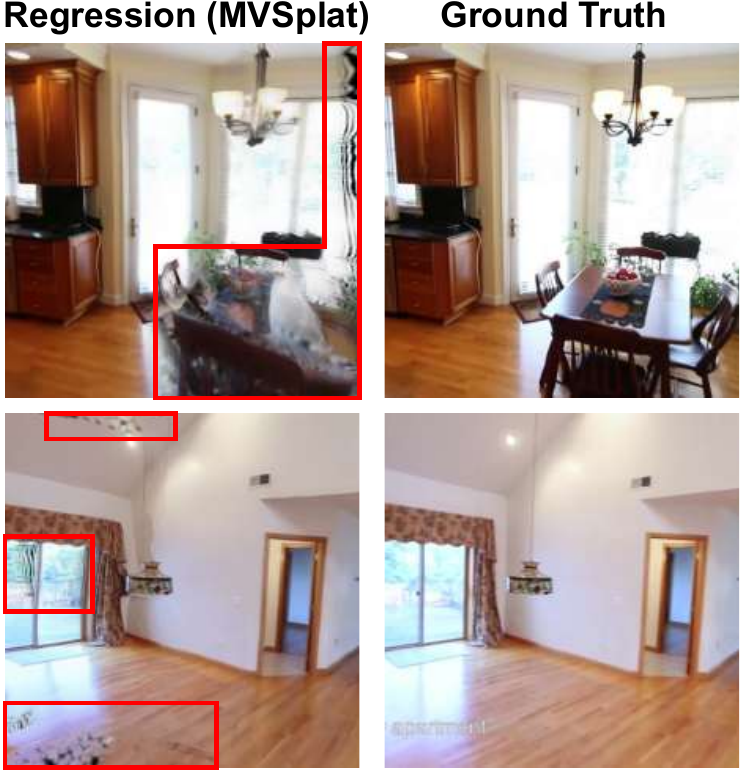}
\vspace{-0.46cm}
\caption{Regression output (MVSplat) and ground truth. Red-highlighted disoccluded regions show artifacts or missing details, while the rest is largely reliable.}
\label{fig:occlusion}
\vspace{-0.47cm}
\end{figure}

To address \textbf{DG3: enable spatially selective refinement}, \THISWORK\ restricts diffusion computation to uncertain or degraded regions while reusing reliable regression output. Regression errors such as disocclusions, motion blur, and low-confidence reconstructions are often spatially localized, as illustrated in Figure~\ref{fig:occlusion}. This creates an opportunity to reduce computation by refining only regions that require correction, provided that the resulting irregular sparsity can be efficiently exploited on GPUs.

\begin{figure}[t]
    \centering
    \includegraphics[width=0.9\linewidth]{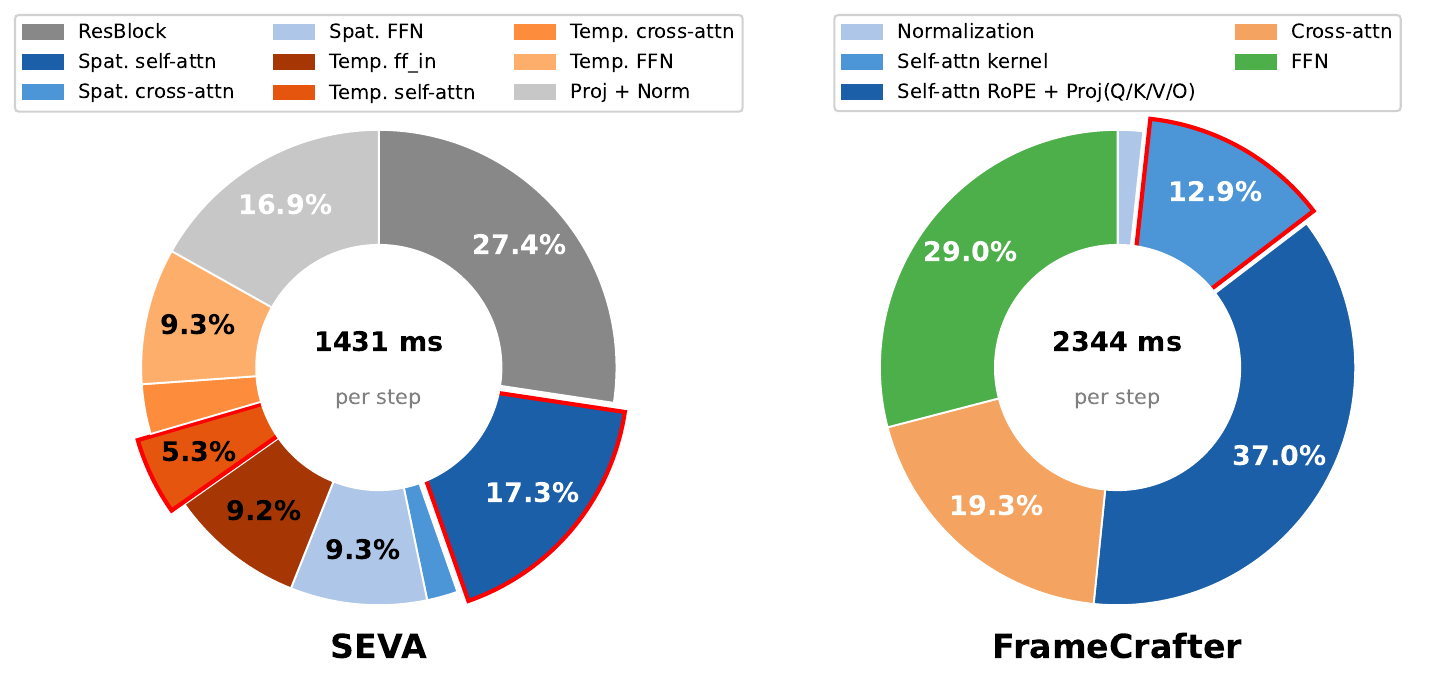}
    \vspace{-0.5cm}
    
    \caption{Per-step latency breakdown of FrameCrafter and SEVA on a single A40. Red boxes mark self-attention components targeted by sparse attention.}
    \vspace{-0.4cm}
    
    \label{fig:diffusion_latency_breakdown}
\end{figure}

Recent diffusion acceleration techniques exploit sparsity or locality in self-attention to reduce computation~\cite{STA, xi2025sparsevideogenacceleratingvideo}. However, in diffusion-based NVS models, self-attention constitutes only a fraction of the per-step denoising cost. As shown in Figure~\ref{fig:diffusion_latency_breakdown}, self-attention accounts for only 12.9\% and 22.6\% of the per-step latency in FrameCrafter and SEVA, respectively, limiting the end-to-end benefit of attention-only sparsification. \THISWORK\ instead exploits spatial sparsity at the refinement level, skipping reliable regions while concentrating diffusion computation on regions that require correction.

Regions for refinement are identified using two complementary cues. Opacity-based confidence uses the per-pixel opacity map from the regression model's 3D Gaussian rasterizer, where low opacity indicates uncertain geometry. Blur-based uncertainty uses a Laplacian-variance metric: regions with low variance are classified as blurry after smoothing, normalization, inversion, and Otsu thresholding~\cite{otsu}. The opacity and blur masks are combined and stacked across frames to form a 3D binary refinement mask.
However, Arbitrary pixel sparsity causes irregular control flow and poor GPU utilization. \THISWORK\ therefore reuses cached latents for unrefined regions and adopts an SBNet-inspired block scheme~\cite{sbnet}: feature maps are tiled, and only blocks containing selected pixels are processed. The selected-token ratio $\sigma_r$ summarizes request $r$'s spatial work.

Regression quality and required diffusion refinement also vary by scene, so \THISWORK\ specializes its refinement policy to scene characteristics.

\subsection{Scene-Aware Performance--Quality Control}

\begin{figure}[t]
    \centering
    \includegraphics[width=0.65\linewidth]{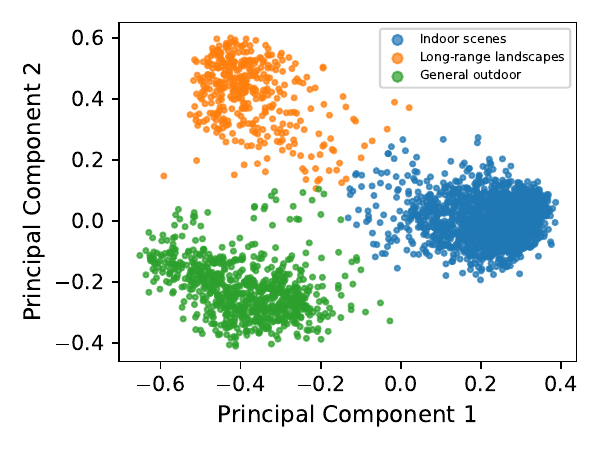}
    \vspace{-0.7cm}
    \caption{Principal Component Analysis (PCA) visualization of input-image embeddings from 3,000 scenes across RE10K, DL3DV, and ACID.}
    \vspace{-0.6cm}
    \label{fig:pca}
\end{figure}

To address \textbf{DG4: provide adaptive quality--performance control across diverse scenes}, \THISWORK\ adapts denoising depth to scene content and camera configuration. Because regression quality varies across scenes, a single global mapping from regression quality to denoising depth can over-refine simple scenes or under-refine challenging ones. Extending Section~\ref{sec:noise_scheduling}, we therefore derive scene-specific $k$-decision logic by clustering CLIP image embeddings~\cite{clip}, with each cluster capturing a distinct quality--efficiency behavior.

As shown in Figure~\ref{fig:pca}, embeddings from 3,000 scenes across RE10K, DL3DV, and ACID form three clusters corresponding to indoor scenes, long-range landscapes, and general outdoor environments. At runtime, \THISWORK\ assigns each input to its nearest cluster and applies its $k$-decision logic, using shallower refinement for simple scenes and deeper denoising for challenging ones. The quality factor $\alpha$ controls denoising depth, while opacity and blur thresholds control the refinement area, together enabling explicit quality--performance trade-offs across diverse scenes.

\subsection{End-to-End Inference Workflow}
\label{subsec:inference_pipeline_overview}


\begin{figure}[t]
    \centering
    \includegraphics[width=\linewidth]{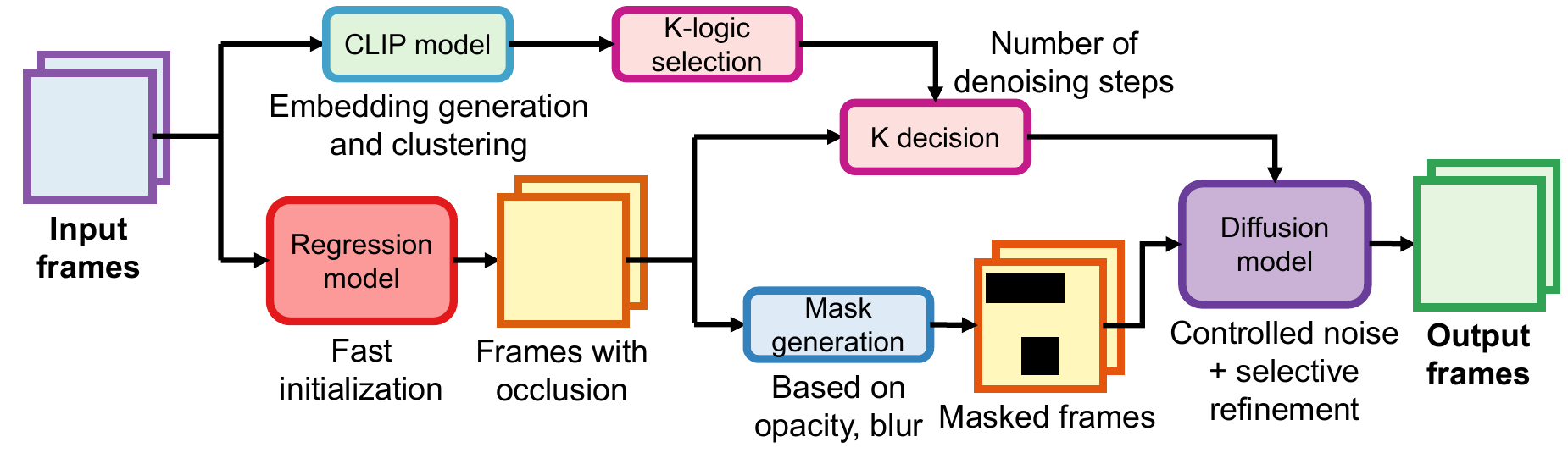}
    \vspace{-0.9cm}
    \caption{Overview of \THISWORK's adaptive inference pipeline.}
    \label{fig:inference_pipeline_overview}
    \vspace{-0.4cm}
\end{figure}

Figure~\ref{fig:inference_pipeline_overview} summarizes the proposed adaptive inference pipeline. Regression first generates coarse target views that provide a structured initialization for diffusion. CLIP-based scene clustering then selects the appropriate $k$-decision logic, while opacity- and blur-based confidence signals identify uncertain regions and define the spatial refinement mask. Diffusion injects controlled noise into the regression outputs and selectively refines only the required frames, timesteps, and spatial regions, while latent reuse preserves temporal consistency during partial denoising. The target quality factor controls denoising depth, whereas the opacity and blur thresholds determine the spatial refinement scope. Together, these mechanisms adapt diffusion computation to each input, reducing redundant work while preserving the desired perceptual quality.

These savings are input-dependent: each request $r$ obtains a denoising depth $k_r$ and refinement mask summarized by selected-token ratio $\sigma_r$, so requests using the same model can have very different remaining diffusion costs. The next section characterizes this inference-generated heterogeneity and its serving implications.

\section{From Adaptive Inference to Request Heterogeneity}
\label{sec:inference_serving}


Section~\ref{sec:inference_design} reduces diffusion computation by
adapting refinement to each regression output. A consequence of this
adaptivity is that requests no longer execute a uniform diffusion
workload. Instead, each request $r$ enters diffusion with a
request-specific denoising depth $k_r$ and selected-token ratio
$\sigma_r$. We next characterize this inference-generated
heterogeneity and explain its implications for online serving.

\subsection{Characterizing Request Heterogeneity}

Adaptive inference introduces request-specific variation along two dimensions. The starting denoising step $k_r$ determines how much of the denoising trajectory is skipped and thus how many denoising steps remain, while the selected-token ratio $\sigma_r$ determines the amount of spatial computation performed within each step. Both depend on the quality and uncertainty of the regression output.

\begin{figure}[t]
    \centering
    \includegraphics[width=\columnwidth]
    {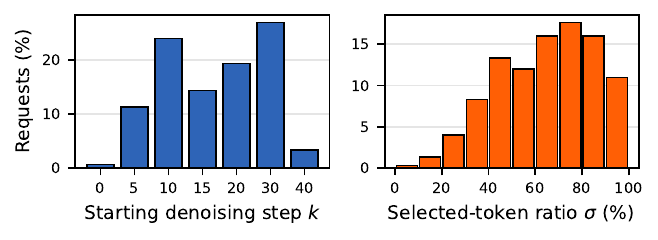}
    \vspace{-1cm}
    \caption{Request heterogeneity in starting denoising steps and selected-token ratios introduced by adaptive inference.}
    \label{fig:heteo_distribution}
    \vspace{-1cm}
\end{figure}

Figure~\ref{fig:heteo_distribution} shows the distribution of $k_r$ and $\sigma_r$ across the 300-request workload, using FrameCrafter as the diffusion model, MVSplat as the regression model, and $\alpha=0.95$. The starting denoising step ranges from 0 to 40, while the selected-token ratio spans from sparse to nearly dense refinement, demonstrating substantial variation in both temporal and spatial computation across requests.

Thus, even requests using the same diffusion model and output resolution can require substantially different amounts of diffusion computation. Requests with larger $k_r$ skip more of the denoising trajectory and execute fewer remaining steps, while requests with larger $\sigma_r$ perform more spatial computation at each step. This inference-generated heterogeneity makes request execution cost input-dependent and motivates a serving system that explicitly accounts for each request's remaining computation.

\subsection{Heterogeneity Under Sequence Parallelism}

\begin{figure}[t]
    \centering
    \includegraphics[width=\columnwidth]
    {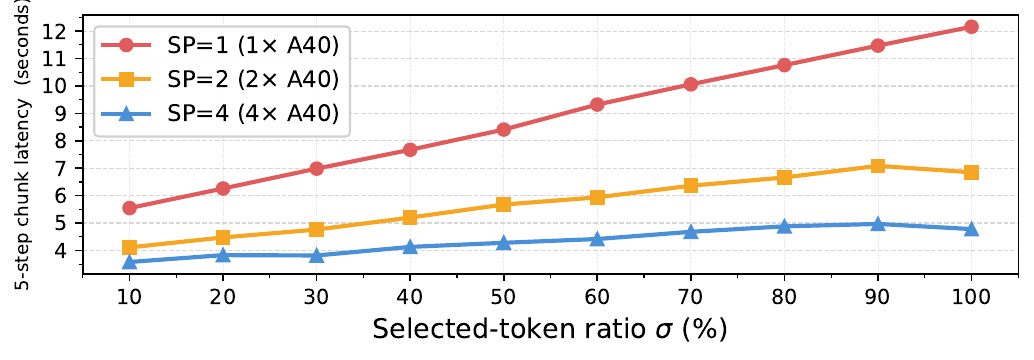}
    \vspace{-1cm}
    \caption{Latency of a five step FrameCrafter denoising chunk on
    four NVIDIA A40 GPUs across token selection ratios and SP degrees.
    Higher token ratios increase latency, while larger SP degrees
    reduce it.}
    \label{fig:latency_sparsity}
    \vspace{-0.5cm}
\end{figure}

Unlike autoregressive LLM inference, whose prefill and decode phases have different computational characteristics, diffusion generation repeatedly executes the denoising network and remains largely compute-bound throughout. Diffusion inference engines such as xDiT~\cite{fang2024xdit} and SwiftFusion~\cite{SwiftFusion} therefore use sequence parallelism (SP) to distribute spatial and temporal tokens across GPUs and reduce latency. With a fixed GPU budget, increasing the SP degree reduces latency but consumes more GPUs per request, limiting concurrency and creating a latency--concurrency trade-off.

\THISWORK\ further complicates this trade-off because adaptive inference introduces request-specific heterogeneity. Even requests using the same model and resolution can have different remaining denoising depths $k_r$ and selected-token ratios $\sigma_r$ after regression, making the execution cost and benefit of additional SP request-dependent. Figure~\ref{fig:latency_sparsity} shows the latency of a five-step FrameCrafter denoising chunk across selected-token ratios and SP degrees. For a fixed SP degree, chunk latency increases with $\sigma_r$ because more spatial tokens participate in denoising. Increasing the SP degree reduces latency, but the benefit depends on $\sigma_r$. Dense requests benefit substantially from additional GPUs, whereas sparse requests may obtain smaller latency reductions.

Together, $k_r$, $\sigma_r$, and the assigned SP degree $p$ determine a request's remaining execution cost: $k_r$ determines the number of remaining denoising chunks, $\sigma_r$ the computation within each chunk, and $p$ how that computation is distributed across GPUs. Our scheduling objective is to maximize SLO attainment by allocating each request the minimum SP degree needed to meet its deadline, while preserving GPU capacity for other requests. Because $k_r$ and $\sigma_r$ vary across requests and evolve during execution, the desired GPU allocation also varies over time.

\subsection{Implications for Serving}

This heterogeneity makes request-oblivious diffusion serving
inefficient. Statically assigning a large SP degree to each request
reduces individual latency but allows a small number of requests to
occupy most GPUs, limiting concurrency. Conversely, uniformly using
a small SP degree preserves concurrency but may leave expensive or
SLO-critical requests unable to complete before their deadlines.

Request ordering must also account for heterogeneous remaining work.
A request with large $k_r$ and $\sigma_r$ can block substantially
lighter requests under First-Come-First-Serve (FCFS), while deadline-only prioritization does
not distinguish requests that require different amounts of computation.
Moreover, because remaining denoising depth decreases after every
denoising round and new requests continuously arrive, the appropriate priority
and SP allocation can change throughout generation.

Efficient serving therefore requires jointly reasoning about
\emph{which requests to execute}, \emph{how many GPUs to allocate to
each request}, and \emph{how much denoising work to execute before
reconsidering these decisions}. This leads to our final design goal,
\textbf{DG5: SLO-aware serving for heterogeneous adaptive workloads}.
Section~\ref{sec:system} describes how \THISWORK\ addresses this goal
by dynamically adapting request priority, SP degree, and chunk size
using the request-specific computation exposed by adaptive inference.

\section{\THISWORK\ Serving System}
\label{sec:system}


Section~\ref{sec:inference_serving} shows that adaptive inference creates request-specific costs in denoising depth and spatial sparsity, while SP trades latency for concurrency. \THISWORK\ therefore schedules at denoising-chunk granularity, jointly adapting request priority, SP degree, chunk size, and GPU placement to remaining work and SLO urgency.

\begin{figure}[t]
    \centering
    \includegraphics[width=0.95\columnwidth]{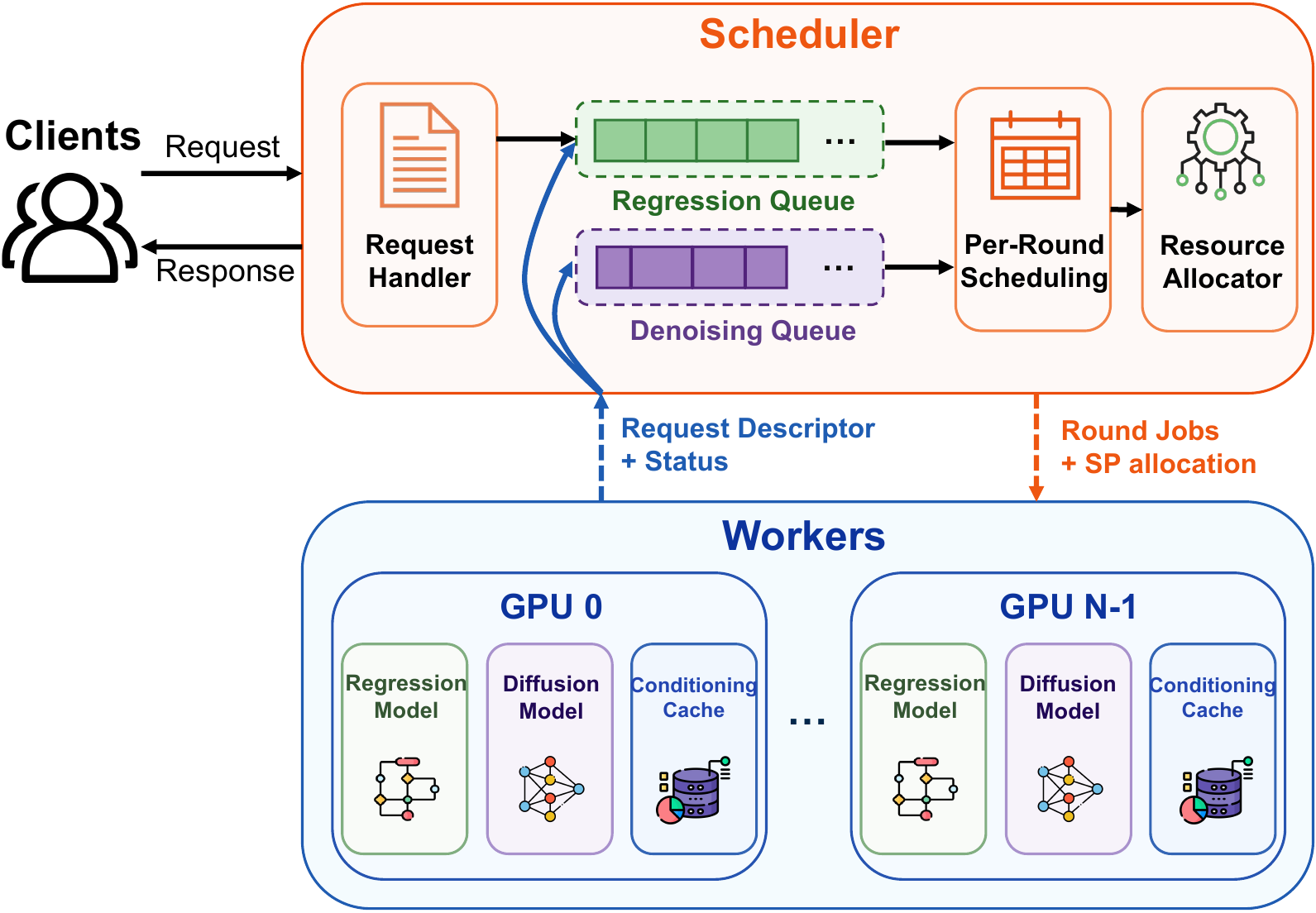}
    \vspace{-0.45cm}
    
    \caption{Overview of the \THISWORK\ serving system.}
    \vspace{-0.5cm}
        
    \label{fig:serving_overview}
\end{figure}

\subsection{Serving Architecture and Request Lifecycle}
\label{sec:serving_overview}

Figure~\ref{fig:serving_overview} shows a centralized CPU scheduler and GPU workers, each executing the adaptive inference pipeline in Figure~\ref{fig:inference_pipeline_overview}. The scheduler maintains regression and denoising queues and tracks each request's deadline, refinement configuration, and progress; workers preload both models and cache request-specific conditioning state generated before diffusion. Requests first run regression, which returns the initialization and descriptor---denoising depth, refinement mask, selected-token ratio, latent state, and reusable conditioning---before entering the denoising queue. Diffusion proceeds in rounds: the scheduler selects requests, SP degrees, GPU groups, and denoising steps, while workers return updated latents and completion status. Unfinished requests are rescheduled at chunk boundaries, enabling reprioritization and SP reallocation without restarting denoising as arrivals, remaining work, and slack change.

\subsection{Request-Aware Cost and Urgency Model}
\label{sec:cost_urgency}

\paragraph{Latency prediction.}
To guide SLO-aware GPU allocation, we propose a profile-based latency predictor that estimates remaining latency from spatial sparsity, denoising location, and SP degree. Let $\mathcal{P}$ denote the supported SP degrees. For request $i$ with selected-token ratio $\sigma_i$, the predicted remaining latency under degree $p$ is
\begin{equation}
    \widehat{T}_i(p)
    =
    \sum_{c\in\mathcal{C}_i}
    \widehat{T}_{\mathrm{chunk}}(p,\sigma_i,k_c)
    +
    \widehat{T}_{\mathrm{post}},
    \label{eq:remaining_latency}
\end{equation}
where $\mathcal{C}_i$ is the set of remaining chunks and $k_c$ is the starting step of chunk $c$. Sparsity values not directly profiled are interpolated, and the estimate is updated after each chunk.

\paragraph{Slack-based prioritization.}
FCFS ignores request cost, while deadline-only scheduling ignores remaining work. \THISWORK\ instead prioritizes requests by slack:
\begin{equation}
    \mathrm{Slack}_i(t)
    =
    (D_i-t)-\widehat{T}_i,
    \label{eq:slack}
\end{equation}
where $D_i$ is the deadline and $\widehat{T}_i$ the predicted remaining execution time. Smaller slack is more urgent and is recomputed each round as requests wait and progress.

\subsection{SLO-Aware Request Selection and SP Allocation}
\label{sec:scheduler}

\begin{algorithm}[t]
\scriptsize
\caption{SLO-Aware Scheduling}
\label{alg:slo_scheduler}
\begin{algorithmic}[1]

\Require Active requests $\mathcal{A}_{\mathrm{reg}},\mathcal{A}_{\mathrm{den}}$,
GPU budget $G$, SP degrees $\mathcal{P}$

\While{requests remain}

\Statex \textcolor{blue}{\textbf{/* Stage 1: Regression initialization */}}
\State $\mathcal{M}\leftarrow
\mathrm{TopG}\!\left(
\mathrm{SortBySlack}(\mathcal{A}_{\mathrm{reg}})
\right)$
\State \Call{Dispatch}{$\mathcal{M}$};
\Call{Wait}{$\mathcal{M}$};
\Call{UpdateConfig}{$\mathcal{M}$}

\Statex \textcolor{blue}{\textbf{/* Stage 2: Build denoising candidates */}}
\State $\mathcal{C}\leftarrow
\mathrm{TopG}\!\left(
\mathrm{SortBySlack}(\mathcal{A}_{\mathrm{den}})
\right)$
\State $\mathcal{F},\mathcal{U}\leftarrow\emptyset,\emptyset$

\ForAll{$r\in\mathcal{C}$}
    \State $\mathcal{P}_r\leftarrow
    \{p\in\mathcal{P}\mid \hat{T}(r,p)\le d_r-t\}$
    \If{$\mathcal{P}_r\neq\emptyset$}
        \State $p_r\leftarrow\min\mathcal{P}_r$;
        $\mathcal{F}\leftarrow\mathcal{F}\cup\{r\}$
    \Else
        \State $\mathcal{U}\leftarrow\mathcal{U}\cup\{r\}$
    \EndIf
\EndFor

\Statex \textcolor{blue}{\textbf{/* Stage 3: Request and GPU allocation */}}
\State $\mathcal{S}\leftarrow
\Call{KnapsackDP}{\mathcal{F},\{p_r\}_{r\in\mathcal{F}},G}$;
$G_{\mathrm{free}}\leftarrow
G-\sum_{r\in\mathcal{S}}p_r$

\ForAll{$r\in
\mathrm{SortBySlack}(\mathcal{F}\setminus\mathcal{S})$}
    \State $\mathcal{P}'_r\leftarrow
    \{p\in\mathcal{P}\mid p\le G_{\mathrm{free}}\}$
    \If{$\mathcal{P}'_r\neq\emptyset$}
        \State $p_r\leftarrow\max\mathcal{P}'_r$;
        $\mathcal{S}\leftarrow\mathcal{S}\cup\{r\}$;
        $G_{\mathrm{free}}\leftarrow G_{\mathrm{free}}-p_r$
    \EndIf
\EndFor

\ForAll{$r\in\mathrm{SortBySlack}(\mathcal{U})$}
    \If{$G_{\mathrm{free}}<\min\mathcal{P}$}
        \State \textbf{break}
    \EndIf
    \State $p_r\leftarrow\min\mathcal{P}$;
    $\mathcal{S}\leftarrow\mathcal{S}\cup\{r\}$;
    $G_{\mathrm{free}}\leftarrow G_{\mathrm{free}}-p_r$
\EndFor

\Statex \textcolor{blue}{\textbf{/* Stage 4: Redistribute GPUs and size chunks */}}
\State $\mathcal{R}\leftarrow
\mathrm{SortBySlack}(\mathcal{S}\cap\mathcal{F})
\Vert
\mathrm{SortBySlack}(\mathcal{S}\cap\mathcal{U})$

\ForAll{$r\in\mathcal{R}$}
    \State $p'_r\leftarrow
    \max\{p\in\mathcal{P}\mid
    p\ge p_r,\; p-p_r\le G_{\mathrm{free}}\}$
    \State $G_{\mathrm{free}}\leftarrow
    G_{\mathrm{free}}-(p'_r-p_r)$;
    $p_r\leftarrow p'_r$
\EndFor

\State $c_r\leftarrow
\hat{T}_{\mathrm{chunk}}(p_r,\sigma_r,k_r),
\ \forall r\in\mathcal{S}$;
$c_{\max}\leftarrow\max_{r\in\mathcal{S}}c_r$

\State $K_r\leftarrow
\min\!\left\{
\left\lfloor K\frac{c_{\max}}{c_r}\right\rfloor,
R_r
\right\},
\ \forall r\in\mathcal{S}$

\State \Call{Dispatch}{$\mathcal{S},\{p_r\},\{K_r\}$};
\Call{Wait}{$\mathcal{S}$};
\Call{UpdateState}{$\mathcal{S}$}

\EndWhile
\end{algorithmic}
\end{algorithm}
To efficiently coordinate regression and denoising while maximizing SLO attainment, we propose the two-phase scheduling procedure in Algorithm~\ref{alg:slo_scheduler}. Phase A runs pending regression requests independently on one GPU each, while Phase B jointly selects denoising requests, SP degrees, and chunk sizes. Phase B scheduling overlaps with Phase A execution.

\paragraph{Minimum feasible SP degree.}
For each candidate request $r$, the scheduler identifies the SP degrees that can complete its remaining work before its deadline:
\begin{equation}
    \mathcal{P}_r
    =
    \left\{
    p\in\mathcal{P}
    \mid
    \widehat{T}(r,p)\leq d_r-t
    \right\}.
    \label{eq:feasible_sp}
\end{equation}
A request is \emph{SLO-feasible} when $\mathcal{P}_r$ is non-empty and receives $p_r=\min\mathcal{P}_r$, preserving GPU capacity while keeping the request on track.

\paragraph{Protecting the maximum number of feasible requests.}
The minimum allocations of all feasible requests may exceed the GPU budget. \THISWORK\ therefore solves
\begin{equation}
    \max_{\mathcal{S}\subseteq\mathcal{F}} |\mathcal{S}|
    \quad
    \mathrm{s.t.}
    \quad
    \sum_{r\in\mathcal{S}} p_r \leq G,
    \label{eq:selection_dp}
\end{equation}
where $\mathcal{F}$ is the feasible set and $G$ the GPU budget. We formulate the selection of SLO-protectable requests as a knapsack problem that maximizes the number of requests meeting their SLOs under the GPU budget. We assign each selected request its minimum feasible SP degree, then consider the remaining feasible requests by increasing slack and allocate the largest supported SP degree that fits. Any leftover capacity serves infeasible requests by urgency at minimum SP to avoid starvation.

\paragraph{Spare-GPU redistribution.}
If GPUs remain idle after scheduling, \textsc{Sphinx}-SAS assigns them to scheduled requests to increase their SP degree and reduce latency. It prioritizes feasible requests before infeasible ones, ordered by slack, without reducing the number of scheduled requests.

\subsection{Adaptive Chunking and Configuration Caching}
\label{sec:adaptive_chunking}

\paragraph{Adaptive chunk size.}
We define the chunk size as the number of denoising steps executed for a request before the scheduler makes the next decision. Because concurrent requests have different per-chunk latencies due to sparsity and SP allocation, a fixed chunk size can leave GPUs idle. Let $K$ be the base chunk size, $c_r=\widehat{T}_{\mathrm{chunk}}(p_r,\sigma_r,k_r)$, and $c_{\max}=\max_{r\in\mathcal{S}}c_r$. The scheduler assigns
\begin{equation}
    K_r
    =
    \min\left(
    \left\lfloor K\frac{c_{\max}}{c_r}\right\rfloor,
    R_r
    \right)
\end{equation}
where $R_r$ is the remaining step count. Faster requests execute more steps per round, approximately aligning completion times and reducing idle GPUs.

\paragraph{Chunk-level preemption.}
After each round, unfinished requests re-enter the scheduling pool with updated slack, allowing new or urgent requests to gain resources at chunk boundaries. SP degree may change with urgency and load; the updated latent state is forwarded to the next SP group without restarting denoising.

\paragraph{Placement-aware worker cache.}
When possible, the scheduler preserves GPU placement so workers can reuse cached image embeddings, camera intrinsic/extrinsic embeddings, and sparse masks. Recomputing this state every round adds over one second of overhead. If placement or SP changes, destination workers generate only missing or SP-dependent state while reusing valid tensors.

\subsection{Scheduling Walk-Through Example}
\begin{figure}[t]
\centering
\includegraphics[width=\columnwidth]{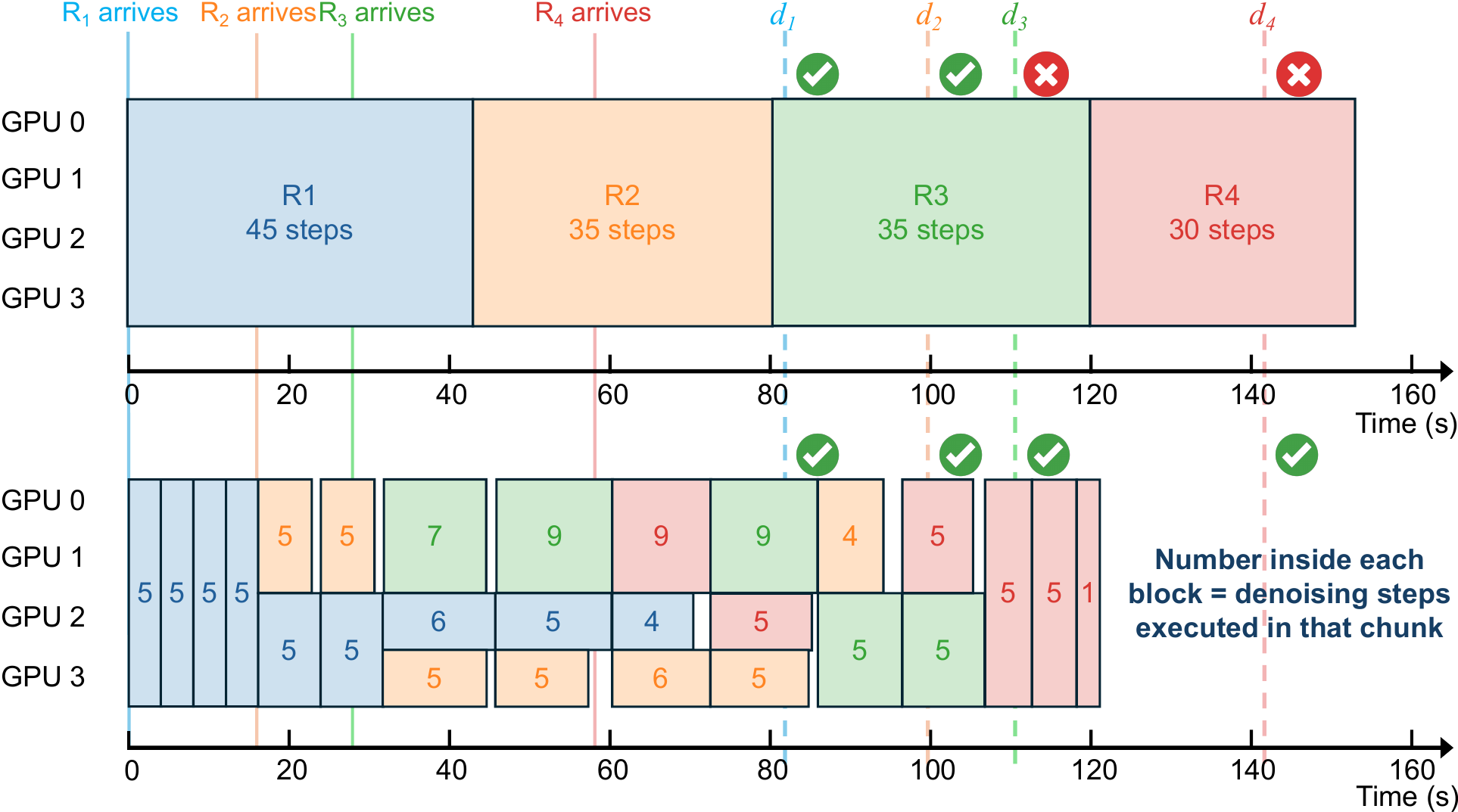}
\vspace{-0.8cm}
\caption{Execution trace comparing xDiT-style static scheduling (top) with
\textsc{Sphinx}-SAS (bottom).}
\vspace{-0.4cm}

\label{fig:scheduling_example}
\end{figure}

Figure~\ref{fig:scheduling_example} shows a real execution trace with four requests that differ in arrival time, denoising depth, and spatial sparsity: $R_1$ (45 denoising steps, 45\% selected tokens), $R_2$ (35, 66\%), $R_3$ (35, 80\%), and $R_4$ (30, 95\%). Under xDiT-style static scheduling, each request monopolizes all four GPUs until completion, minimizing the latency of the currently running request but blocking later arrivals; the resulting queueing delay causes $R_3$ and $R_4$ to miss their deadlines. In contrast, \textsc{Sphinx}-SAS (\textbf{S}LO-\textbf{A}ware \textbf{S}cheduler) reschedules active requests at every chunk boundary using updated remaining cost and SLO slack. It admits multiple requests for concurrent execution, adjusts their SP degrees as GPU capacity changes, and varies chunk sizes to better align round completion times and reduce idle resources. As a result, GPU capacity shifts dynamically toward more urgent requests instead of remaining fixed for an entire generation, allowing all four requests to complete within their deadlines.

\section{Evaluation Methodology} \label{section:experimental_setup}

\noindent\textbf{\underline{Datasets and Models.}} 
We evaluate \THISWORK\ on three representative NVS benchmarks: RealEstate10K (RE10K)~\cite{re10k}, ACID~\cite{acid}, and DL3DV~\cite{dl3dv}. 
Our hybrid pipeline integrates regression- and diffusion-based models. For diffusion, we use SEVA~\cite{seva} and FrameCrafter~\cite{Wu2026framecrafter}; for regression, we adopt MVSplat~\cite{mvsplat} for both SEVA and FrameCrafter. 
Image quality is evaluated using standard metrics, including PSNR~\cite{psnr}, SSIM~\cite{ssim}, LPIPS~\cite{lpips}, and MUSIQ~\cite{musiq}.

\noindent \textbf{\underline{Baselines.}}
To assess the effectiveness of our hybrid design, we compare against the following baselines:

\begin{itemize}[leftmargin=*]
    \setlength{\itemsep}{1pt}
    \setlength{\parskip}{0pt}
    \setlength{\parsep}{0pt}

    \item \textbf{Diffusion-Only}: Full diffusion-based generation without regression initialization.
    
    \item \textbf{Regression-Only}: Deterministic regression output without diffusion-based refinement.
    \item \textbf{STA}~\cite{STA}: Sliding Tile Attention, which replaces full attention with tile-based local attention to exploit spatiotemporal locality and reduce attention computation.
    \item \textbf{\THISWORK}: The proposed two-stage framework combining regression-based initialization with selective denoising.
\end{itemize}






\noindent\textbf{\underline{Implementation Details.}}
We implement \THISWORK, our serving system, in Python using PyTorch~\cite{ansel2024pytorch}. We evaluate \THISWORK\ on two hardware configurations: a server equipped with four NVIDIA A40 GPUs, each with 48GB of VRAM, and a server equipped with eight NVIDIA H200 GPUs, each with 141GB of VRAM.
We use a total of 50 diffusion steps for all models, following each model's default configuration. During online serving, we use a base chunk size of 5 denoising steps on A40 GPUs and 10 steps on H200 GPUs.
For selective refinement, we apply opacity- and blur-based confidence masks to identify regions that require diffusion-based refinement. The input and target resolutions follow the default settings of each model: 576$\times$576 for SEVA and 480$\times$832 for FrameCrafter. SEVA takes 2 input frames as context and generates 19 target frames between them, while FrameCrafter takes 3 input frames and generates 1 target frame, following their respective default configurations. In terms of STA, we implement model-specific Triton~\cite{Triton} kernels for each model: for SEVA, we use a $4\times4\times4$ tile geometry (temporal$\times$height$\times$width) yielding 64 tokens per tile with a $5\times3\times3$ sliding window (45 tiles); for FrameCrafter, we use a $4\times2\times4$ tile geometry yielding 32 tokens per tile with a $7\times5$ sliding tile window. Energy consumption is measured using Zeus~\cite{Zeus}.

\noindent \textbf{\underline{Experimental Setup.}}
We conduct both online and offline experiments to comprehensively evaluate \THISWORK. For online serving, we measure tail latency and SLO attainment under dynamic workloads. For offline evaluation, we measure average inference latency and output image quality.

\textbf{\textit{Latency and Image Quality Experiments:}}
Offline experiments use 1,000 requests per dataset on an A40 and report average end-to-end inference latency and image quality.

\textbf{\textit{Online Serving Experiments:}}
We evaluate FrameCrafter at $\alpha=0.95$ using 300 requests, 100 per dataset, and compare four strategies: (1) \textbf{Diffusion-Only}, full diffusion with FCFS; (2) \textbf{Sphinx-NonSparse}, partial denoising with FCFS; (3) \textbf{Sphinx-Sparse}, partial and selective denoising with FCFS; and (4) \textbf{Sphinx-SAS}, Sphinx-Sparse with our SLO-aware scheduler. Requests arrive as a Poisson process at varying rates. We use SLO scales of $1.5\times$ and $2.0\times$, relative to each request's offline Vanilla latency at the fastest SP degree.

\section{Results}

\subsection{Inference Efficiency}

\begin{figure*}[t]
    \centering
    \includegraphics[width=\linewidth]{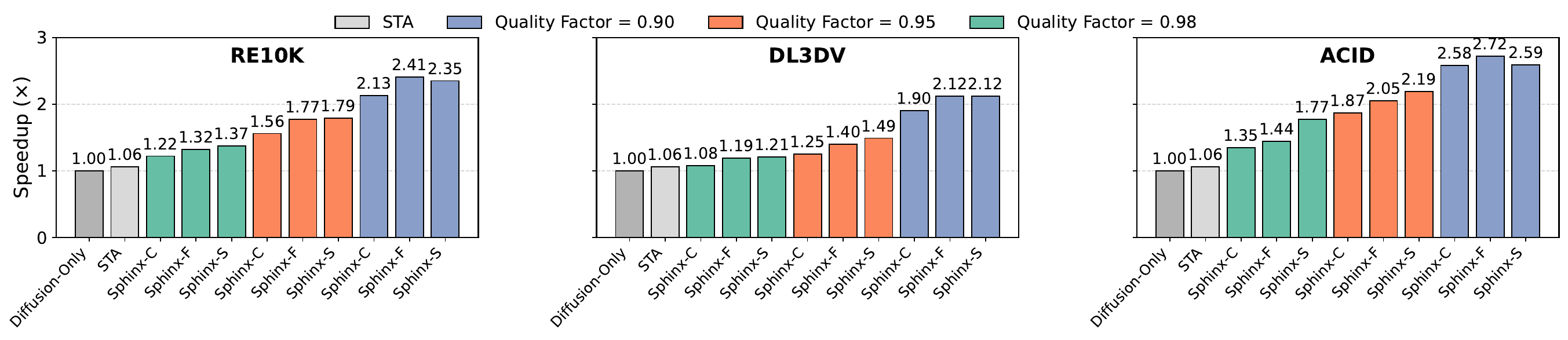}
    \vspace{-1cm}
    \caption{Comparison of SEVA performance across RE10K, DL3DV, and ACID under different quality settings. Each bar represents a refinement mode (Fine, Coarse, Selective), and colors correspond to quality factors 0.98, 0.95, and 0.90.}
    \vspace{-0.5cm}
    
    \label{fig:speedup_seva}
\end{figure*}

\begin{figure*}[t]
    \centering
    \includegraphics[width=\linewidth]{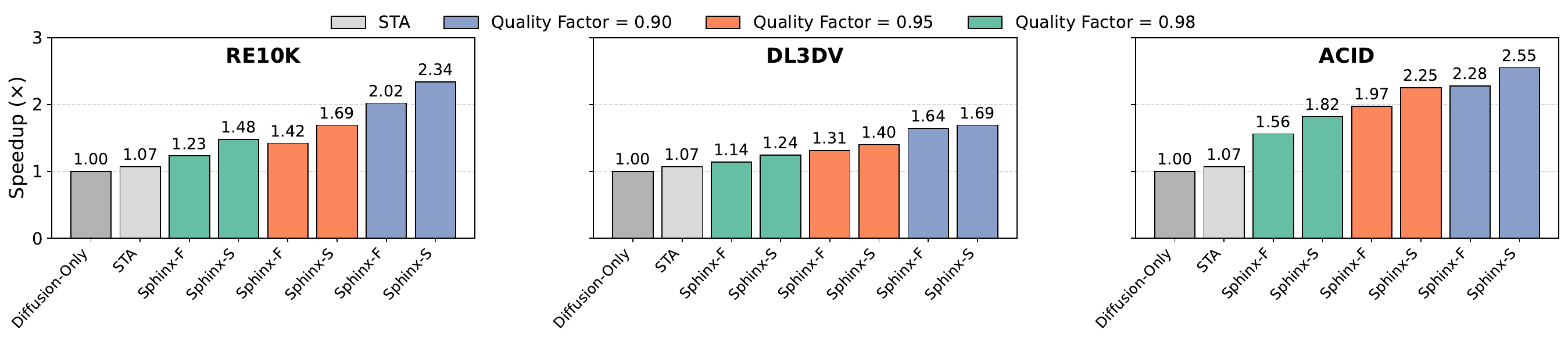}
    \vspace{-1cm}
    \caption{Comparison of FrameCrafter performance across RE10K, DL3DV, and ACID under different quality settings. Each bar represents a refinement mode (Fine, Coarse, Selective), and colors correspond to quality factors 0.98, 0.95, and 0.90.}
    \vspace{-0.55cm}
    
    \label{fig:speedup_framecrafter}
\end{figure*}

\begin{figure}[t]
    \centering
    \includegraphics[width=0.9\linewidth]{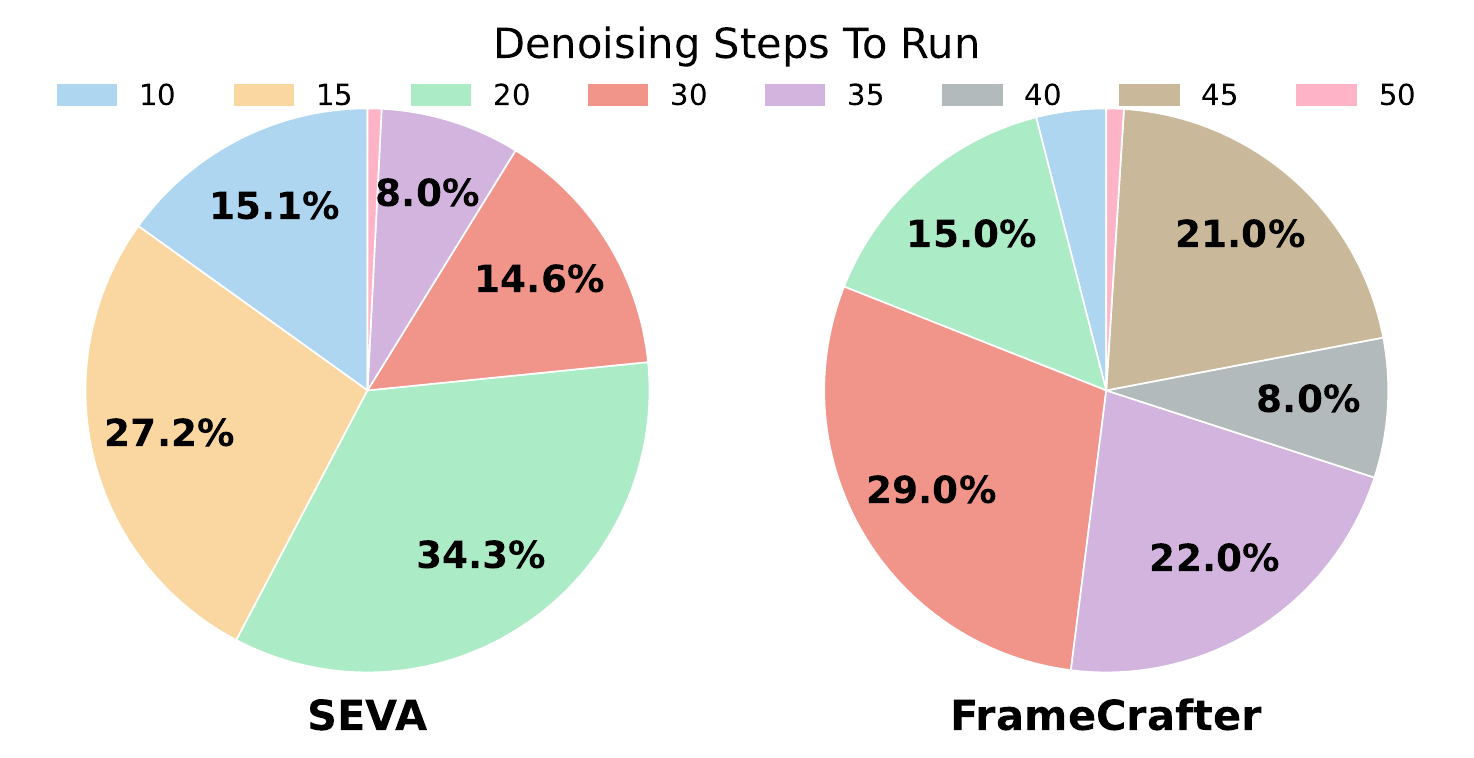}
    \vspace{-0.5cm}
    \caption{Distribution of target frames requiring different numbers of denoising steps for SEVA and FrameCrafter on the RE10K dataset at a quality factor of 0.95.}
    \vspace{-0.4cm}
    
    \label{fig:steps_distribution}
\end{figure}
Figure~\ref{fig:speedup_seva} compares the performance of \THISWORK\ on SEVA across RE10K, DL3DV, and ACID under three refinement modes and different quality factors. {\THISWORK-F} uses latent caching and reuse to enable fine-grained temporal adaptivity, allowing each frame to be denoised only for the timesteps it requires. {\THISWORK-C} disables latent caching and applies the maximum predicted denoising depth to all target frames, while {\THISWORK-S} further skips refinement in high-confidence regions. Compared with full diffusion inference, \THISWORK\ substantially reduces end-to-end latency across all datasets, achieving up to $1.8\times$ speedup on ACID at $\alpha=0.98$ and over $2\times$ at $\alpha=0.90$. These results demonstrate \textbf{DG1: adaptively applying denoising to balance quality and efficiency}. 

Within the same quality factor, {\THISWORK-F} consistently outperforms {\THISWORK-C} by avoiding unnecessary refinement, while {\THISWORK-S} provides additional gains at $\alpha=0.98$ and $0.95$ by exploiting spatial sparsity.
The benefit of selective refinement is moderate because the U-Net architecture progressively downsamples latent features at deeper layers. To preserve selected regions, the confidence mask is max-pooled at each layer, reducing effective sparsity at lower resolutions. At $\alpha=0.90$, fewer denoising steps remain, so mask generation and sparse scatter--gather overhead can offset the saved computation, leading to limited additional speedup or a slight slowdown relative to {\THISWORK-F}. Figure~\ref{fig:speedup_framecrafter} shows a similar trend for FrameCrafter. Since it generates only one target frame per request, {\THISWORK-C} is not applicable. \THISWORK\ still consistently reduces latency, achieving over $1.6\times$ speedup at $\alpha=0.95$ on RE10K. The smaller gain compared with SEVA is likely due to lower-quality regression initialization, which requires more diffusion steps to reach the target quality.

We further compare \THISWORK\ with STA, which provides only 6--7\% end-to-end acceleration because self-attention accounts for only 12.9\% and 22.6\% of FrameCrafter and SEVA per-step latency, respectively, as shown in Figure~\ref{fig:diffusion_latency_breakdown}. In contrast, \THISWORK\ reduces computation across the full denoising trajectory. Figure~\ref{fig:steps_distribution} shows that at $\alpha=0.95$, more FrameCrafter samples require 30 or more denoising steps, while most SEVA frames require fewer, confirming that stronger regression initialization enables larger speedups.

\subsection{Quality Evaluation}
Table~\ref{tab:seva_quality} evaluates \THISWORK\ on SEVA across RE10K, DL3DV, and ACID using MUSIQ, PSNR, SSIM, and LPIPS. At $\alpha=0.98$, results remain close to Diffusion-Only, typically within 1--2\%, indicating that temporal adaptivity and latent reuse introduce little quality degradation. As $\alpha$ decreases to 0.95 and 0.90, quality gradually declines while remaining within the target range, showing that partial denoising can reduce computation while preserving perceptual quality. In contrast, STA degrades substantially across all four metrics, suggesting that aggressively reducing attention computation without quality-aware adaptation harms generation quality.

These results demonstrate \textbf{DG2: temporal coherence under partial denoising}. Although target frames may execute different numbers of denoising steps, latent reuse maintains cross-frame consistency while avoiding redundant computation. Selective refinement introduces only a small additional quality loss, with limited changes in metrics such as MUSIQ and SSIM, further supporting \textbf{DG3: spatially selective refinement} by concentrating diffusion computation on uncertain regions while reusing reliable regression outputs.

Table~\ref{tab:framecrafter_quality} shows similar trends for FrameCrafter. However, DL3DV exhibits a larger quality drop with selective refinement at $\alpha=0.95$ and $0.90$, indicating that selective refinement should be disabled when regression quality is insufficient for reliable reuse. Overall, \THISWORK\ provides a flexible quality--efficiency trade-off across models and datasets.
\begin{table}[!t]
\centering
\footnotesize
\setlength{\tabcolsep}{4pt}
\renewcommand{\arraystretch}{1.15}
\resizebox{\columnwidth}{!}{%
\begin{tabular}{llcccc}
\toprule
\multicolumn{2}{c}{\textbf{SEVA}} & \multicolumn{4}{c}{\textbf{Quality Metrics}} \\
\cmidrule(lr){1-2}\cmidrule(lr){3-6}
\textbf{Benchmark} & \textbf{Method} &
\textbf{MUSIQ}$\uparrow$ &
\textbf{PSNR}$\uparrow$ &
\textbf{SSIM}$\uparrow$ &
\textbf{LPIPS}$\downarrow$ \\
\midrule

\multirow{9}{*}{\textbf{RE10K}}
 & Diffusion-Only
 & 63.55 & 21.28 & 0.736 & 0.250 \\

 & \cellcolor{gray!10}STA
 & \cellcolor{gray!10}55.27
 & \cellcolor{gray!10}12.84
 & \cellcolor{gray!10}0.472
 & \cellcolor{gray!10}0.495 \\

 & \THISWORK\ ($\alpha = 0.98$)
 & 63.40 & 21.14 & 0.737 & 0.250 \\

 & \cellcolor{gray!10}\THISWORK\ ($\alpha = 0.98$) + selective refinement
 & \cellcolor{gray!10}62.53
 & \cellcolor{gray!10}21.00
 & \cellcolor{gray!10}0.737
 & \cellcolor{gray!10}0.257 \\

 & \THISWORK\ ($\alpha = 0.95$)
 & 62.46 & 20.63 & 0.730 & 0.262 \\

 & \cellcolor{gray!10}\THISWORK\ ($\alpha = 0.95$) + selective refinement
 & \cellcolor{gray!10}62.10
 & \cellcolor{gray!10}20.61
 & \cellcolor{gray!10}0.730
 & \cellcolor{gray!10}0.265 \\

 & \THISWORK\ ($\alpha = 0.90$)
 & 60.23 & 20.21 & 0.724 & 0.279 \\

 & \cellcolor{gray!10}\THISWORK\ ($\alpha = 0.90$) + selective refinement
 & \cellcolor{gray!10}60.06
 & \cellcolor{gray!10}20.20
 & \cellcolor{gray!10}0.725
 & \cellcolor{gray!10}0.280 \\

 & Regression-Only
 & 55.93 & 20.01 & 0.726 & 0.304 \\
\midrule

\multirow{9}{*}{\textbf{DL3DV}}
 & Diffusion-Only
 & 65.43 & 17.98 & 0.468 & 0.379 \\

 & \cellcolor{gray!10}STA
 & \cellcolor{gray!10}59.27
 & \cellcolor{gray!10}12.84
 & \cellcolor{gray!10}0.292
 & \cellcolor{gray!10}0.555 \\

 & \THISWORK\ ($\alpha = 0.98$)
 & 63.85 & 17.12 & 0.451 & 0.393 \\

 & \cellcolor{gray!10}\THISWORK\ ($\alpha = 0.98$) + selective refinement
 & \cellcolor{gray!10}64.39
 & \cellcolor{gray!10}17.13
 & \cellcolor{gray!10}0.449
 & \cellcolor{gray!10}0.398 \\

 & \THISWORK\ ($\alpha = 0.95$)
 & 62.96 & 16.49 & 0.439 & 0.407 \\

 & \cellcolor{gray!10}\THISWORK\ ($\alpha = 0.95$) + selective refinement
 & \cellcolor{gray!10}62.86
 & \cellcolor{gray!10}16.17
 & \cellcolor{gray!10}0.431
 & \cellcolor{gray!10}0.422 \\

 & \THISWORK\ ($\alpha = 0.90$)
 & 59.96 & 15.42 & 0.408 & 0.455 \\

 & \cellcolor{gray!10}\THISWORK\ ($\alpha = 0.90$) + selective refinement
 & \cellcolor{gray!10}60.39
 & \cellcolor{gray!10}15.43
 & \cellcolor{gray!10}0.407
 & \cellcolor{gray!10}0.460 \\

 & Regression-Only
 & 56.59 & 14.87 & 0.369 & 0.517 \\
\midrule

\multirow{9}{*}{\textbf{ACID}}
 & Diffusion-Only
 & 64.79 & 23.86 & 0.643 & 0.260 \\

 & \cellcolor{gray!10}STA
 & \cellcolor{gray!10}62.01
 & \cellcolor{gray!10}17.54
 & \cellcolor{gray!10}0.487
 & \cellcolor{gray!10}0.393 \\

 & \THISWORK\ ($\alpha = 0.98$)
 & 64.85 & 23.93 & 0.651 & 0.263 \\

 & \cellcolor{gray!10}\THISWORK\ ($\alpha = 0.98$) + selective refinement
 & \cellcolor{gray!10}63.57
 & \cellcolor{gray!10}23.87
 & \cellcolor{gray!10}0.657
 & \cellcolor{gray!10}0.273 \\

 & \THISWORK\ ($\alpha = 0.95$)
 & 64.17 & 23.79 & 0.654 & 0.268 \\

 & \cellcolor{gray!10}\THISWORK\ ($\alpha = 0.95$) + selective refinement
 & \cellcolor{gray!10}63.39
 & \cellcolor{gray!10}23.72
 & \cellcolor{gray!10}0.655
 & \cellcolor{gray!10}0.274 \\

 & \THISWORK\ ($\alpha = 0.90$)
 & 63.27 & 23.62 & 0.654 & 0.275 \\

 & \cellcolor{gray!10}\THISWORK\ ($\alpha = 0.90$) + selective refinement
 & \cellcolor{gray!10}63.17
 & \cellcolor{gray!10}23.60
 & \cellcolor{gray!10}0.654
 & \cellcolor{gray!10}0.276 \\

 & Regression-Only
 & 62.60 & 24.60 & 0.730 & 0.247 \\
\bottomrule
\end{tabular}%
}
\caption{SEVA quality results on RE10K, DL3DV, and ACID. Higher is better except for LPIPS.}
\vspace{-0.5cm}
\label{tab:seva_quality}
\end{table}

\begin{table}[!t]
\centering
\footnotesize
\setlength{\tabcolsep}{4pt}
\renewcommand{\arraystretch}{1.15}
\resizebox{\columnwidth}{!}{%
\begin{tabular}{llcccc}
\toprule
\multicolumn{2}{c}{\textbf{FrameCrafter}} & \multicolumn{4}{c}{\textbf{Quality Metrics}} \\
\cmidrule(lr){1-2}\cmidrule(lr){3-6}
\textbf{Benchmark} & \textbf{Method} &
\textbf{MUSIQ}$\uparrow$ &
\textbf{PSNR}$\uparrow$ &
\textbf{SSIM}$\uparrow$ &
\textbf{LPIPS}$\downarrow$ \\
\midrule

\multirow{9}{*}{\textbf{RE10K}}
 & Diffusion-Only
 & 58.99 & 19.39 & 0.640 & 0.193 \\

 & \cellcolor{gray!10}STA
 & \cellcolor{gray!10}57.13
 & \cellcolor{gray!10}14.51
 & \cellcolor{gray!10}0.476
 & \cellcolor{gray!10}0.439 \\

 & \THISWORK\ ($\alpha = 0.98$)
 & 57.51 & 19.28 & 0.644 & 0.210 \\

 & \cellcolor{gray!10}\THISWORK\ ($\alpha = 0.98$) + selective refinement
 & \cellcolor{gray!10}55.03
 & \cellcolor{gray!10}19.09
 & \cellcolor{gray!10}0.650
 & \cellcolor{gray!10}0.235 \\

 & \THISWORK\ ($\alpha = 0.95$)
 & 56.45 & 19.09 & 0.648 & 0.228 \\

 & \cellcolor{gray!10}\THISWORK\ ($\alpha = 0.95$) + selective refinement
 & \cellcolor{gray!10}54.54
 & \cellcolor{gray!10}18.95
 & \cellcolor{gray!10}0.653
 & \cellcolor{gray!10}0.245 \\

 & \THISWORK\ ($\alpha = 0.90$)
 & 55.64 & 17.45 & 0.617 & 0.306 \\

 & \cellcolor{gray!10}\THISWORK\ ($\alpha = 0.90$) + selective refinement
 & \cellcolor{gray!10}55.00
 & \cellcolor{gray!10}17.75
 & \cellcolor{gray!10}0.629
 & \cellcolor{gray!10}0.292 \\

 & Regression-Only
 & 56.98 & 18.47 & 0.650 & 0.276 \\
\midrule

\multirow{9}{*}{\textbf{DL3DV}}
 & Diffusion-Only
 & 68.32 & 16.79 & 0.438 & 0.258 \\

 & \cellcolor{gray!10}STA
 & \cellcolor{gray!10}63.71
 & \cellcolor{gray!10}12.46
 & \cellcolor{gray!10}0.294
 & \cellcolor{gray!10}0.540 \\

 & \THISWORK\ ($\alpha = 0.98$)
 & 67.13 & 16.41 & 0.428 & 0.284 \\

 & \cellcolor{gray!10}\THISWORK\ ($\alpha = 0.98$) + selective refinement
 & \cellcolor{gray!10}65.35
 & \cellcolor{gray!10}16.14
 & \cellcolor{gray!10}0.424
 & \cellcolor{gray!10}0.315 \\

 & \THISWORK\ ($\alpha = 0.95$)
 & 65.43 & 15.73 & 0.411 & 0.328 \\

 & \cellcolor{gray!10}\THISWORK\ ($\alpha = 0.95$) + selective refinement
 & \cellcolor{gray!10}59.37
 & \cellcolor{gray!10}13.62
 & \cellcolor{gray!10}0.357
 & \cellcolor{gray!10}0.488 \\

 & \THISWORK\ ($\alpha = 0.90$)
 & 61.52 & 14.52 & 0.384 & 0.416 \\

 & \cellcolor{gray!10}\THISWORK\ ($\alpha = 0.90$) + selective refinement
 & \cellcolor{gray!10}57.26
 & \cellcolor{gray!10}13.21
 & \cellcolor{gray!10}0.352
 & \cellcolor{gray!10}0.513 \\

 & Regression-Only
 & 56.89 & 14.03 & 0.369 & 0.496 \\
\midrule

\multirow{9}{*}{\textbf{ACID}}
 & Diffusion-Only
 & 63.53 & 20.56 & 0.497 & 0.253 \\

 & \cellcolor{gray!10}STA
 & \cellcolor{gray!10}57.98
 & \cellcolor{gray!10}14.81
 & \cellcolor{gray!10}0.364
 & \cellcolor{gray!10}0.539 \\

 & \THISWORK\ ($\alpha = 0.98$)
 & 61.81 & 20.88 & 0.531 & 0.279 \\

 & \cellcolor{gray!10}\THISWORK\ ($\alpha = 0.98$) + selective refinement
 & \cellcolor{gray!10}61.27
 & \cellcolor{gray!10}21.10
 & \cellcolor{gray!10}0.541
 & \cellcolor{gray!10}0.282 \\

 & \THISWORK\ ($\alpha = 0.95$)
 & 60.62 & 20.89 & 0.540 & 0.296 \\

 & \cellcolor{gray!10}\THISWORK\ ($\alpha = 0.95$) + selective refinement
 & \cellcolor{gray!10}60.48
 & \cellcolor{gray!10}21.01
 & \cellcolor{gray!10}0.546
 & \cellcolor{gray!10}0.295 \\

 & \THISWORK\ ($\alpha = 0.90$)
 & 59.88 & 20.74 & 0.542 & 0.308 \\

 & \cellcolor{gray!10}\THISWORK\ ($\alpha = 0.90$) + selective refinement
 & \cellcolor{gray!10}59.81
 & \cellcolor{gray!10}20.87
 & \cellcolor{gray!10}0.546
 & \cellcolor{gray!10}0.304 \\

 & Regression-Only
 & 62.46 & 21.18 & 0.561 & 0.273 \\
\bottomrule
\end{tabular}%
}
\caption{FrameCrafter quality results on RE10K, DL3DV, and ACID. Higher is better except for LPIPS.}
\vspace{-0.7cm}

\label{tab:framecrafter_quality}
\end{table}

\subsection{SLO Compliance and Tail Latency}

\begin{figure}[t]
    \centering
    \includegraphics[width=\linewidth]{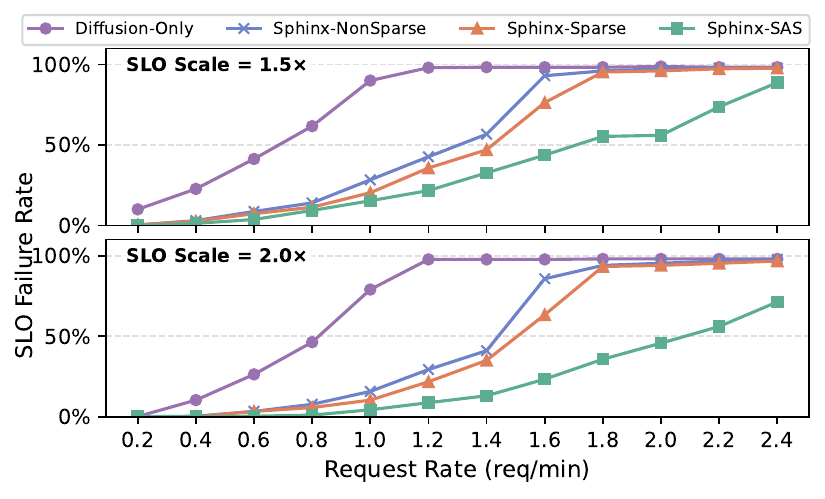}
    \vspace{-1cm}
    \caption{SLO failure rate under varying request rates and SLO scales on a server with 4 NVIDIA A40 GPUs.}
    \vspace{-0.3cm}
    
    \label{fig:slo_a40}
\end{figure}

\begin{figure}[t]
    \centering
    \includegraphics[width=\linewidth]{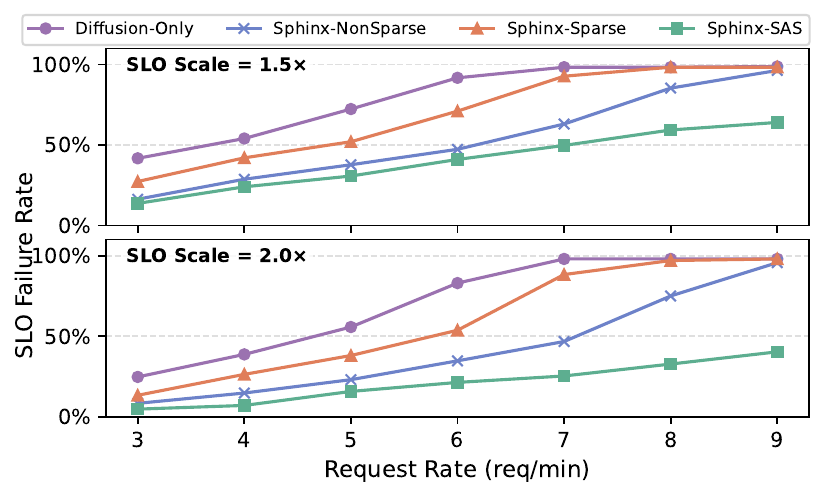}
    \vspace{-1cm}
    \caption{SLO failure rate under varying request rates and SLO scales on a server with 8 NVIDIA H200 GPUs.}
    \vspace{-0.5cm}
    
    \label{fig:slo_h200}
\end{figure}

\begin{figure}[t]
    \centering
    \includegraphics[width=\linewidth]{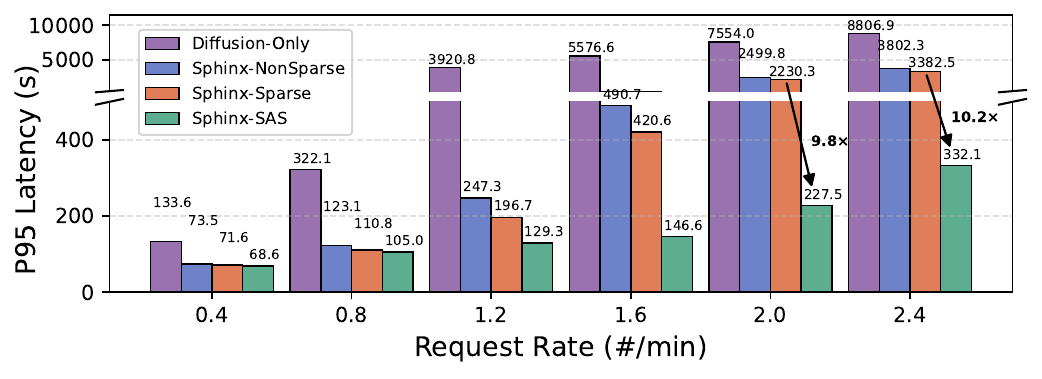}
    \vspace{-1cm}
    \caption{P95 tail latency of serving strategies across request rates on 4 NVIDIA A40 GPUs.}
    \vspace{-0.5cm}
    
    \label{fig:p95_a40}
\end{figure}

\begin{figure}[t]
    \centering
    \includegraphics[width=\linewidth]{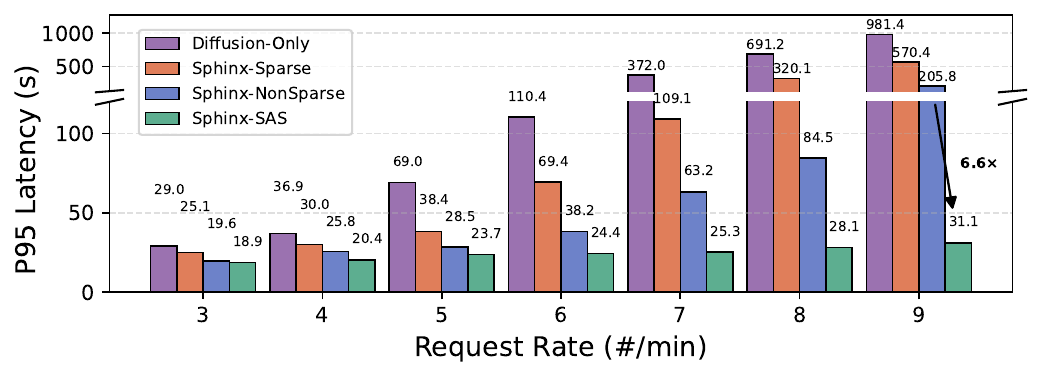}
    \vspace{-1cm}
    \caption{P95 tail latency of serving strategies across request rates on 8 NVIDIA H200 GPUs.}
    \vspace{-0.5cm}

    \label{fig:p95_h200}
\end{figure}

\begin{figure}[t]
    \centering
    \includegraphics[width=\linewidth]{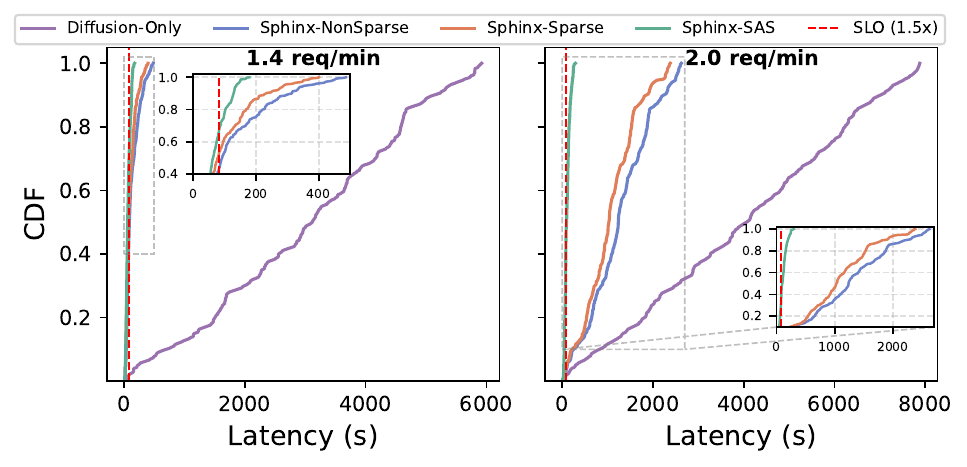}
    \vspace{-1cm}
    \caption{CDF of end-to-end latency for different serving strategies under request rates of 1.4 and 2.0 requests/min on a server with 4 NVIDIA A40 GPUs. The red dashed line indicates the SLO threshold under an SLO scale of $1.5\times$.}
    \vspace{-0.3cm}
    
    \label{fig:cdf_a40}
\end{figure}

\begin{figure}[t]
    \centering
    \includegraphics[width=\linewidth]{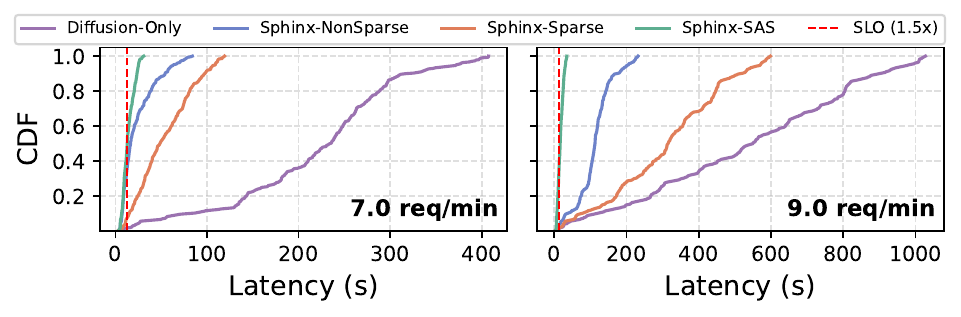}
    \vspace{-0.9cm}
    \caption{CDF of end-to-end latency for different serving strategies under request rates of 7.0 and 9.0 requests/min on a server with 8 NVIDIA H200 GPUs. The red dashed line indicates the SLO threshold under an SLO scale of $1.5\times$.}
    \vspace{-0.4cm}
    \label{fig:cdf_h200}
\end{figure}

Figures~\ref{fig:slo_a40} and~\ref{fig:slo_h200} compare SLO failure rates as request rate increases. On A40, reducing inference computation consistently improves SLO attainment: Sphinx-NonSparse substantially outperforms Diffusion-only, selective refinement in Sphinx-Sparse provides additional gains, and \textsc{Sphinx}-SAS achieves the lowest failure rate across the load range. For example, at 1.4 requests/min under the $1.5\times$ SLO, the failure rate drops from nearly 100\% for Diffusion-only to approximately 57\% for Sphinx-NonSparse, 47\% for Sphinx-Sparse, and 33\% for \textsc{Sphinx}-SAS. More importantly, at 1.8 requests/min, \textsc{Sphinx}-SAS achieves about 45\% SLO attainment versus only 5\% for Sphinx-Sparse, a nearly $9\times$ improvement despite using the same adaptive inference. This gap highlights the importance of SLO-aware scheduling: reallocating GPUs according to request urgency and remaining computation prevents heterogeneous requests from saturating the system under high load. On H200, Sphinx-NonSparse outperforms Sphinx-Sparse because sparse generation is slower than full generation in most configurations. H200's high compute capability and large sequence-parallelism degrees substantially reduce dense denoising latency, while mask generation and sparse-execution overheads remain, causing selective refinement costs to outweigh its computation savings. \textsc{Sphinx}-SAS avoids this inefficiency by using offline latency profiles to choose between sparse and full generation for each request and SP degree. When full generation with a large SP degree is predicted faster, it selects full generation instead. The relative improvement of \textsc{Sphinx}-SAS is therefore smaller than on A40, since denoising chunks execute much faster on H200 and scheduling and preemption overheads become more noticeable. Nevertheless, \textsc{Sphinx}-SAS remains the best-performing strategy, especially under high load; at 9 requests/min, it maintains failure rates of approximately 65\% and 40\% under the $1.5\times$ and $2.0\times$ SLOs, respectively, while the other strategies approach 95--100\% failure.

Figures~\ref{fig:p95_a40} and~\ref{fig:p95_h200} show that the latency benefits of \textsc{Sphinx}-SAS become more pronounced under heavy load. On A40, its P95 latency reaches only 227.5s and 332.1s at 2.0 and 2.4 requests/min, respectively, corresponding to $9.8\times$ and $10.2\times$ reductions over Sphinx-Sparse. On H200, \textsc{Sphinx}-SAS maintains a consistently stable P95 latency as the request rate increases, growing only from around 19,s at 3 requests/min to 31.1,s at 9 requests/min, while the FCFS baselines increase rapidly under high load. At 9 requests/min, its P95 latency is $6.6\times$ lower than the next-best Sphinx-NonSparse baseline. These results show that SLO-aware scheduling is particularly effective near saturation, preventing queueing delays from rapidly increasing tail latency.
Figures~\ref{fig:cdf_a40} and~\ref{fig:cdf_h200} show the same trend in the full latency distribution. On A40, \textsc{Sphinx}-SAS keeps latency concentrated within a few hundred seconds even at 2.0 requests/min, while the FCFS baselines develop tails extending to thousands of seconds. On H200, \textsc{Sphinx}-SAS similarly maintains the most concentrated distribution at both 7 and 9 requests/min. Overall, adaptive inference improves serving capacity, while SLO-aware scheduling further controls tail latency as the system approaches saturation. Together, these results demonstrate \textbf{DG5: SLO-aware serving for heterogeneous adaptive workloads}, showing that dynamically adapting resource allocation to request urgency and remaining computation improves SLO attainment and tail latency under high load.

\subsection{Navigating Quality--Efficiency Trade-offs}
\begin{figure}[t]
    \centering
    \includegraphics[width=\linewidth]{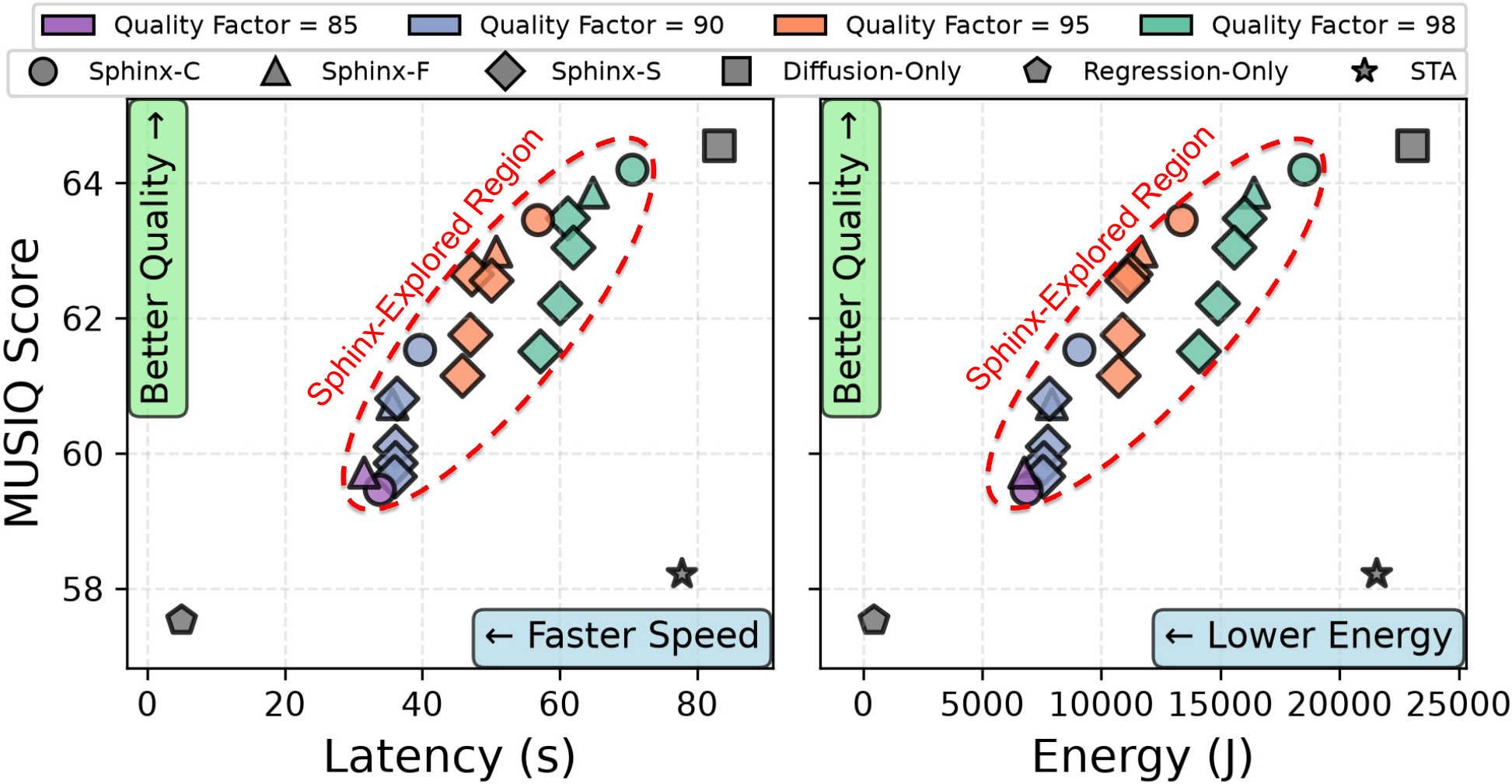}
    \vspace{-0.9cm}
    \caption{Performance--quality and energy--quality trade-offs across quality factors and inference methods.}
    \vspace{-0.4cm}
    \label{fig:trade_off}
\end{figure}

Figure~\ref{fig:trade_off} summarizes the trade-offs among perceptual quality, latency, and energy using SEVA across RE10K, DL3DV, and ACID. The left figure plots average latency against MUSIQ, while the right figure plots energy consumption against MUSIQ; lower latency and energy and higher perceptual quality are desirable. Across four quality factors (0.85--0.98) and three refinement modes, \THISWORK-C, \THISWORK-F, and \THISWORK-S, \THISWORK\ spans a broad operating region between the fast, low-energy Regression-Only baseline and the higher-quality but more expensive Diffusion-Only baseline. Increasing the quality factor allocates more diffusion computation, improving perceptual quality at the cost of higher latency and energy, while selective refinement provides additional opportunities to reduce unnecessary computation. Overall, these results demonstrate \textbf{DG4: adaptive quality--performance control across diverse scenes}: by adjusting refinement depth and spatial selectivity, \THISWORK\ provides multiple operating points along both the quality--latency and quality--energy trade-offs, allowing deployments to balance perceptual fidelity, latency, and energy according to runtime constraints.

\section{Related Work}

Sparse VideoGen~\cite{xi2025sparsevideogenacceleratingvideo}, FastVideo~\cite{STA}, Hash3D~\cite{hash3d}, Li~\textit{et al.}~\cite{li2022efficient}, and VideoScene~\cite{wang2025videoscenedistillingvideodiffusion} reduce diffusion cost via sparsity, caching, distillation, or sparse execution, while \THISWORK\ adapts denoising depth and spatial refinement. BiLD~\cite{BiLD}, SpecInfer~\cite{specinfer}, Apparate~\cite{Apparate}, and E$^3$~\cite{E3} exploit lightweight models or per-input computation. LLM serving systems including LoongServe~\cite{LoongServe}, Sarathi-Serve~\cite{agrawal2024taming}, Llumnix~\cite{Llumnix}, NanoFlow~\cite{NanoFlow}, DistServe~\cite{DistServe}, AdaServe~\cite{AdaServe}, TokenFlow~\cite{TokenFlow}, and adaptive pipeline parallelism~\cite{adaptivePP} use elastic parallelism, chunking, placement, preemption, and SLO-aware allocation. Diffusion serving systems MoDM~\cite{modm}, Argus~\cite{Argus}, MixFusion~\cite{mixfusion}, TetriServe~\cite{tetriserve2026}, and GENSERVE~\cite{genserve} address model, configuration, resolution, or workload heterogeneity, whereas \THISWORK\ targets regression-induced heterogeneity in denoising steps and selected-token ratios. M*~\cite{mstar2026} provides a multimodal runtime, while VibeServe~\cite{VibeServe} synthesizes serving systems without scheduling request-dependent states.

\section{Conclusions}
We introduced \textbf{\THISWORK}, an NVS inference and serving framework combining regression initialization with diffusion refinement. \THISWORK\ reduces diffusion computation through adaptive partial denoising and spatially selective refinement, while its SLO-aware scheduler adapts request priority, SP degree, chunk size, and sparse/dense execution. Across two models and three datasets, \THISWORK\ achieves a $2.3\times$ average speedup (up to $2.7\times$) with less than 10\% perceptual degradation. Tested on A40s and H200s, it also improves SLO attainment and reduces tail latency under load.

\section{Acknowledgments}
We used AI tools for code development, figure plotting, and text editing. This work was supported in part by the Semiconductor Research Corporation (SRC) under the GRC AIHW grant, the AMD AI \& HPC Cluster Program, and Advanced Micro Devices, Inc. through the Funding Academic Research program, NSF grants CCF-2450085 and CNS-2106184, and DARPAML2P Award HR0011-26-9-E190. This work used the Delta advanced computing resource at the National Center for Supercomputing Applications (NCSA) through the National Science Foundation's Advanced Cyberinfrastructure Coordination Ecosystem: Services \& Support (ACCESS) program under allocation CIS261050.

\balance
\bibliography{99_ref.bib}
\bibliographystyle{unsrt} 

\clearpage
\appendix
\onecolumn
\section*{Appendix}

\section{Algorithmic Overview of Sphinx Inference}
\label{sec:sphinx_algo}
This section provides a detailed walkthrough of the Sphinx inference pipeline summarized in Algorithm~\ref{alg:sphinx}. Sphinx combines regression-based fast initialization with adaptive diffusion refinement to achieve a controllable quality--efficiency trade-off for novel view synthesis. Given input views and camera poses, Sphinx first applies a lightweight regression model to generate coarse target frames and per-pixel opacity estimates, which serve both as initialization for diffusion and as indicators of uncertain regions (Line 1). The input scene is then encoded using a CLIP encoder and assigned to a pre-defined cluster, from which a cluster-specific denoising-step decision logic (\emph{k}-logic) is selected to determine the minimum refinement depth required to satisfy a user-specified quality target (Lines 2--6). To avoid unnecessary computation on confident regions, Sphinx constructs spatial refinement masks by combining a Laplacian-variance–based uncertainty metric with opacity-based confidence, yielding a binary mask that identifies regions requiring refinement (Lines 7--10). The regression outputs are subsequently encoded into the diffusion latent space, and noise is injected according to the selected denoising depth, initializing the diffusion process at a reduced timestep rather than from pure noise (Lines 11--12). During inference, Sphinx periodically performs full denoising steps to refresh temporal attention states, while intermediate steps selectively refine only masked regions and active frames; cached temporal representations are reused for inactive frames to preserve cross-frame consistency and eliminate redundant computation (Lines 13--22). After completing the scheduled diffusion steps, the final latent representations are decoded to produce the output images (Lines 23--24).

\begin{algorithm}[h] \caption{Sphinx Inference} \label{alg:sphinx} \begin{adjustbox}{max width=\columnwidth} \begin{minipage}{\columnwidth} \small \begin{algorithmic}[1] \Require Inputs $I=\{I_0,I_1\}$ with poses; target pose batch $\Pi=\{\pi_t\}_{t=1}^{N}$; quality $\alpha$; opacity threshold $\tau_o$; cache period $T$; total steps $S$ \Statex \textbf{Models:} Regression $R$; Diffusion $D$; CLIP encoder $E$; MUSIQ score $\mathcal{Q}$ \Ensure Frames $\mathbf{Y}\in\mathbb{R}^{N\times H\times W\times C}$ \Statex \textcolor{blue}{\textbf{// Fast init using Regression model}} \State $(\mathbf{X},\,\mathbf{O}) \gets R(I,\Pi)$ \Comment{regression outputs: coarse frames and opacity} \Statex \textcolor{blue}{\textbf{// K-logic selection}} \State $e \gets E(I)$;\quad $c \gets \mathrm{cluster}(e)$;\quad $\mathcal{K}_c \gets \mathrm{k\_logic}(c,\alpha)$ \Statex \textcolor{blue}{\textbf{// Starting step selection}} \State $\hat{\mathbf{Q}} \gets \mathcal{Q}(\mathbf{X})$ \Comment{MUSIQ per item} \State $Q^\star(t) \gets \mathrm{interp}\!\big(\mathcal{Q}(I_0),\mathcal{Q}(I_1), t/N\big)$ \State $\mathbf{r} \gets \hat{\mathbf{Q}} \oslash Q^\star$ \Comment{per-item ratio} \State $\mathbf{k} \gets \mathcal{K}_c(\mathbf{r})$;\quad $k_{\min} \gets \min(\mathbf{k})$ \Statex \textcolor{blue}{\textbf{// Spatial masks generation}} \State $\mathbf{B} \gets \mathrm{laplacian\_var}(\mathbf{X})$ \State $\mathbf{M}^{\text{blur}} \gets \mathrm{otsu}(\mathrm{norm}(\mathbf{B}))$ \Comment{binary mask from Otsu} \State $\mathbf{M}^{\text{op}} \gets \mathbf{1}[\mathbf{O} < \tau_o]$ \State $\mathbf{M} \gets \mathbf{M}^{\text{op}} \lor \mathbf{M}^{\text{blur}}$ \Statex \textcolor{blue}{\textbf{// Latent initialization \& noise injection}} \State $\mathbf{Z}^{(0)} \gets \mathrm{encode}(\mathbf{X})$ \State $\mathbf{Z}^{(k_{\min})} \gets \mathrm{add\_noise}(\mathbf{Z}^{(0)},\,k_{\min})$ \Comment{noisy latents at start step} \Statex \textcolor{blue}{\textbf{// Partial denoising using Diffusion model}} \For{$u = k_{\min}$ \textbf{to} $S-1$} \If{$u \bmod T = 0$} \State $\mathbf{Z}^{(u+1)} \gets D.\mathrm{full\_step}(\mathbf{Z}^{(u)},\,u)$; \State $\mathrm{cache} \gets \mathrm{store\_temporal}()$ \Else \State $\mathbf{A}_u \gets \mathbf{1}[\mathbf{k} \le u]$ \Comment{frames active at step $u$} \State $\mathbf{Z}^{(u+1)}_{\mathbf{A}_u} \gets D.\mathrm{partial\_step}(\mathbf{Z}^{(u)}_{\mathbf{A}_u},\,u,\,\mathbf{M}_{\mathbf{A}_u},\,\mathrm{cache})$ \State $\mathbf{Z}^{(u+1)}_{\overline{\mathbf{A}}_u} \gets \mathrm{add\_noise}\!\left(\mathbf{Z}^{(0)}_{\overline{\mathbf{A}}_u},\,u{+}1\right)$ \Comment{inactive frames: resample to step $u{+}1$ from clean latents} \EndIf \EndFor \Statex \textcolor{blue}{\textbf{// Decode latent outputs to final images}} \State $\mathbf{Y} \gets \mathrm{decode}(\mathbf{Z}^{(S)})$ \State \Return $\mathbf{Y}$ \end{algorithmic} \end{minipage} \end{adjustbox} \end{algorithm}

\section{Latency Breakdown}
\label{sec:latency_breakdown}

\begin{figure}[H]
    \centering
    \includegraphics[width=0.7\linewidth]{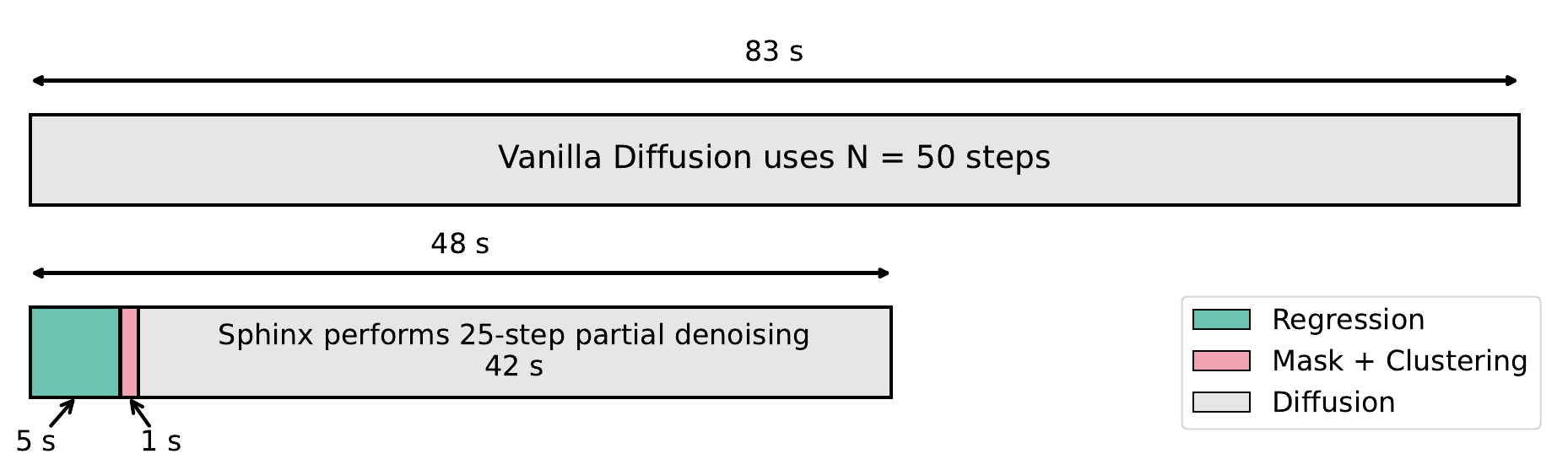}
    \vspace{-0.4cm}
    \caption{End-to-end latency breakdown of vanilla diffusion and \textsc{Sphinx}.}
    \label{fig:sphinx_timing_breakdown}
\end{figure}

To understand where runtime is spent and how our hybrid pipeline reduces end-to-end latency, Figure~\ref{fig:sphinx_timing_breakdown} compares the inference-time breakdown of vanilla diffusion and \textsc{Sphinx}. Vanilla diffusion executes the full $N{=}50$-step denoising trajectory, resulting in an end-to-end latency of 83\,s. In contrast, \textsc{Sphinx} starts from a fast regression-based initialization and performs only 25 denoising steps. This reduces the diffusion stage to 42\,s and the total latency to 48\,s, corresponding to a 1.7$\times$ speedup over vanilla diffusion.

The additional overhead introduced by \textsc{Sphinx} is modest: regression initialization takes 5\,s, while mask generation and scene clustering together take 1\,s. This 6\,s overhead is substantially outweighed by the reduction in diffusion computation from 83\,s to 42\,s. Overall, the breakdown shows that diffusion remains the dominant source of latency and that \textsc{Sphinx} achieves most of its savings by reducing the number of denoising steps.

\section{Visual Examples}
\label{sec:visual_examples}

\begin{figure*}[h]
    \centering
    \renewcommand{\arraystretch}{1.2}
    \setlength{\tabcolsep}{2pt} 
    \small
    \begin{tabular}{c c c c c c}
        \hline
         \textbf{Input Frame 1} & \textbf{Regression (MVSplat)} & \textbf{\THISWORK\ ($\alpha=0.95$)} & \textbf{Diffusion (SEVA)} & \textbf{Input Frame 2} \\
        \hline

        \includegraphics[width=0.16\textwidth]{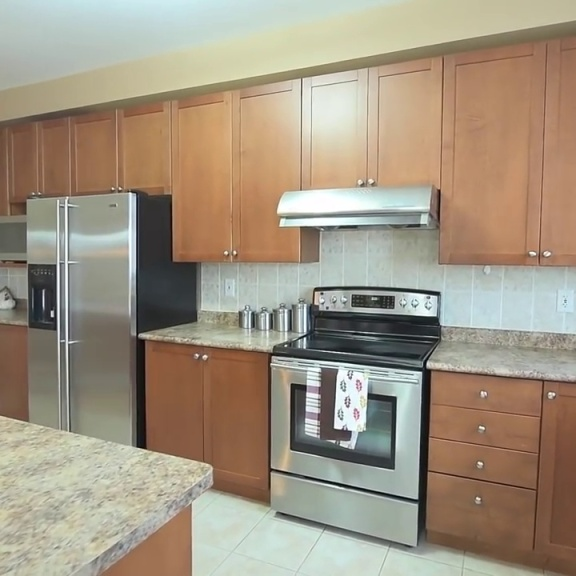} & 
        \includegraphics[width=0.16\textwidth]{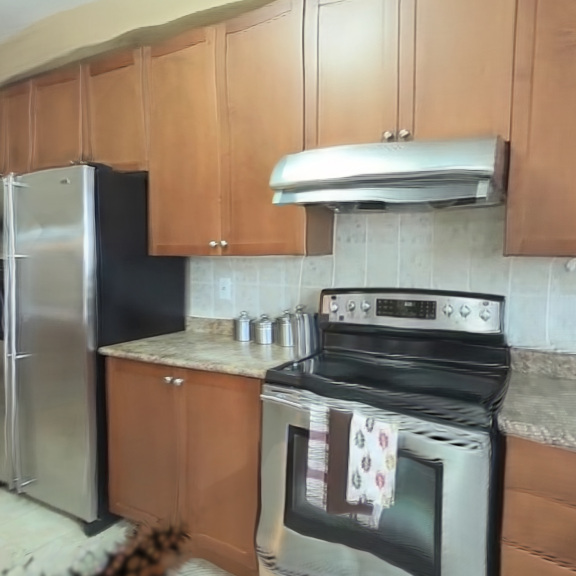} & 
        \includegraphics[width=0.16\textwidth]{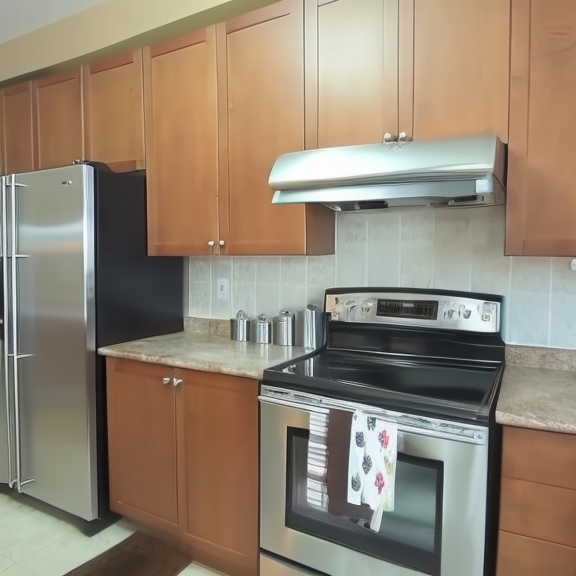} &
        \includegraphics[width=0.16\textwidth]{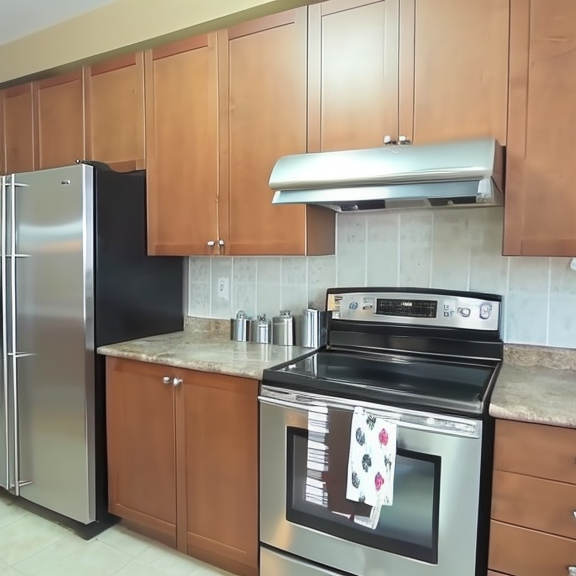} & 
        \includegraphics[width=0.16\textwidth]{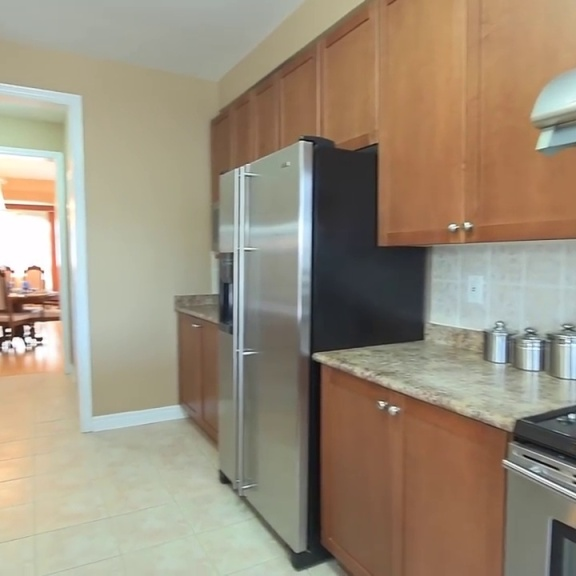} \\

        \includegraphics[width=0.16\textwidth]{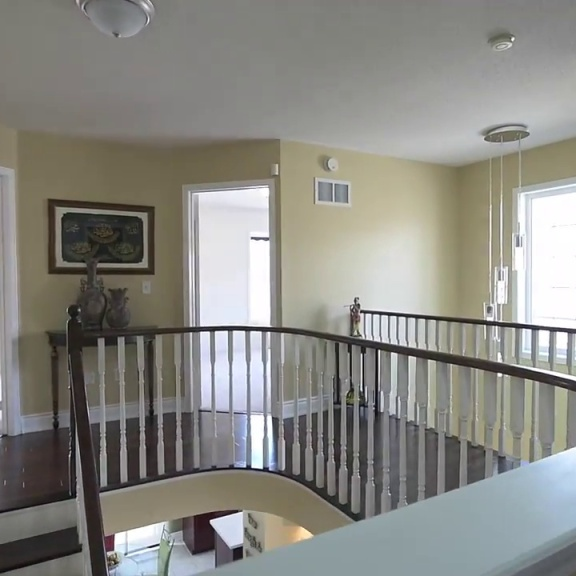} & 
        \includegraphics[width=0.16\textwidth]{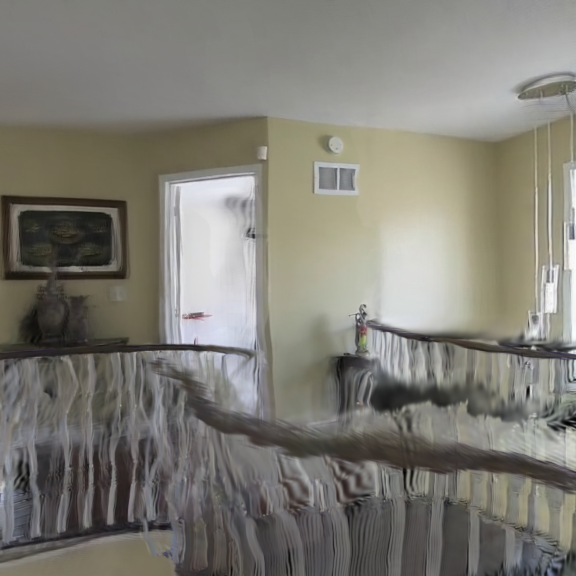} & 
        \includegraphics[width=0.16\textwidth]{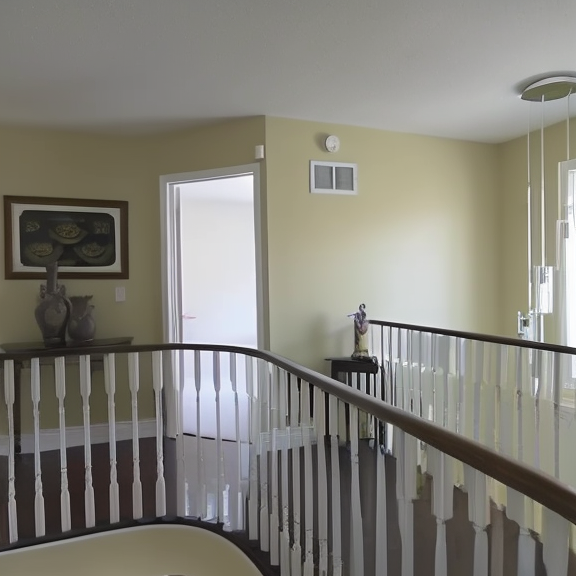} &
        \includegraphics[width=0.16\textwidth]{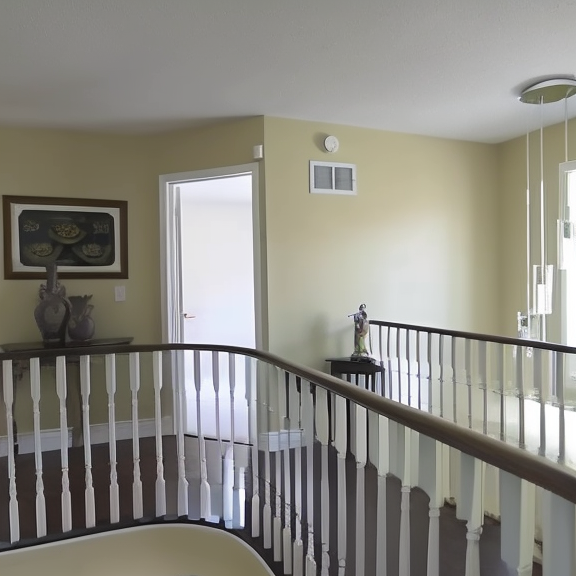} & 
        \includegraphics[width=0.16\textwidth]{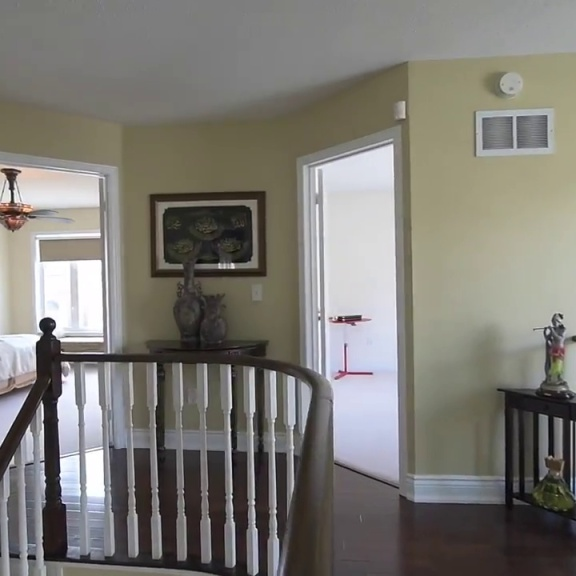} \\

        \includegraphics[width=0.16\textwidth]{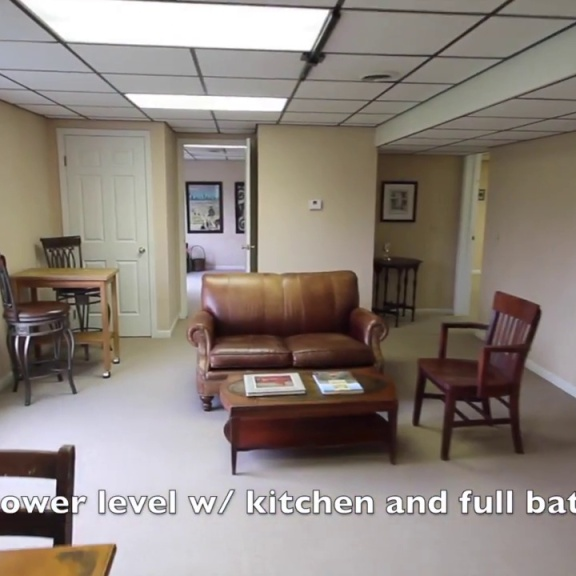} & 
        \includegraphics[width=0.16\textwidth]{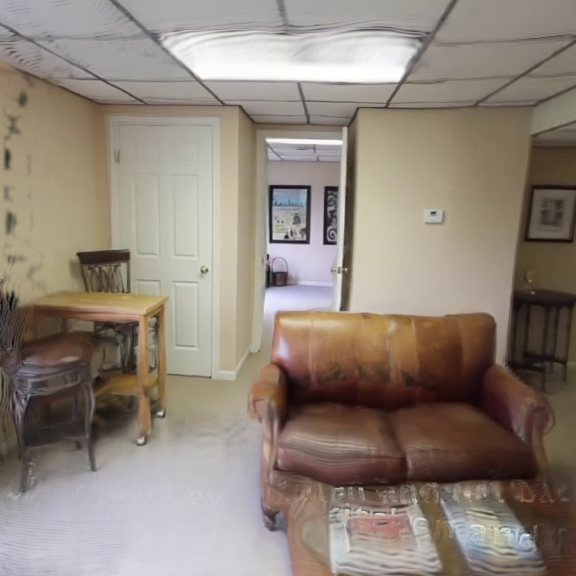} & 
        \includegraphics[width=0.16\textwidth]{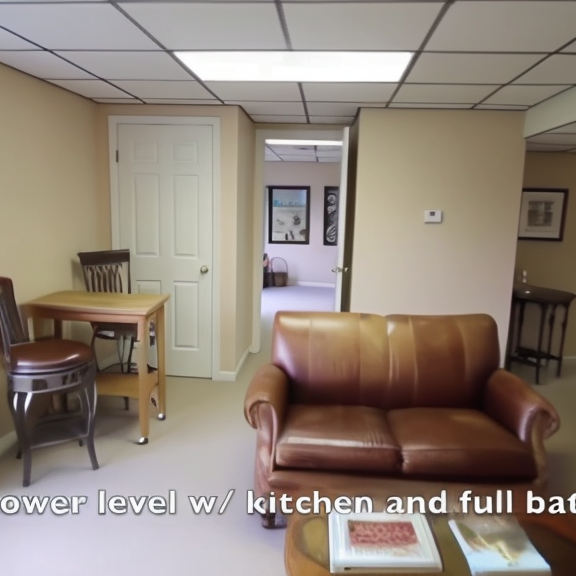} &
        \includegraphics[width=0.16\textwidth]{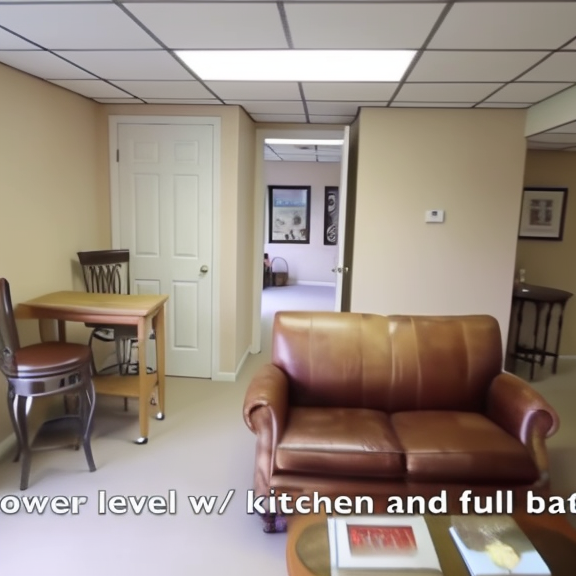} & 
        \includegraphics[width=0.16\textwidth]{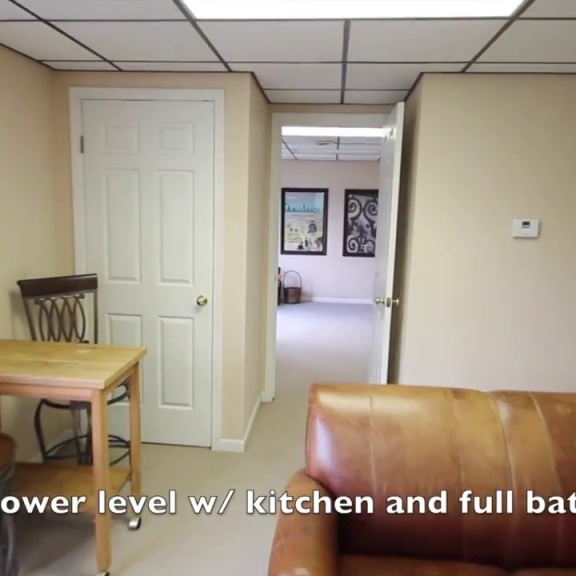} \\

        \includegraphics[width=0.16\textwidth]{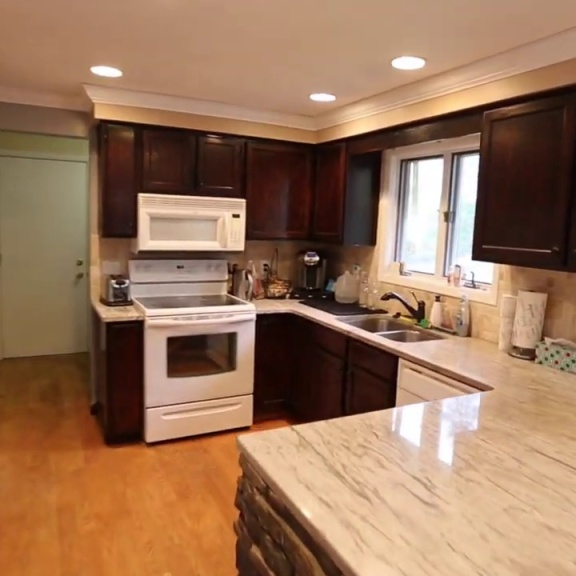} & 
        \includegraphics[width=0.16\textwidth]{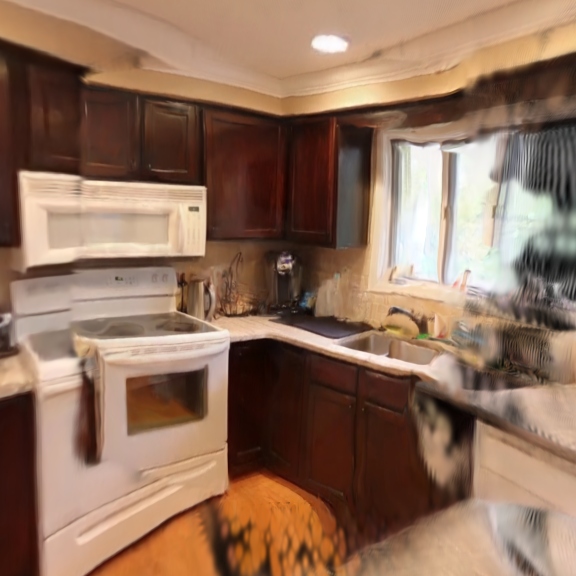} & 
        \includegraphics[width=0.16\textwidth]{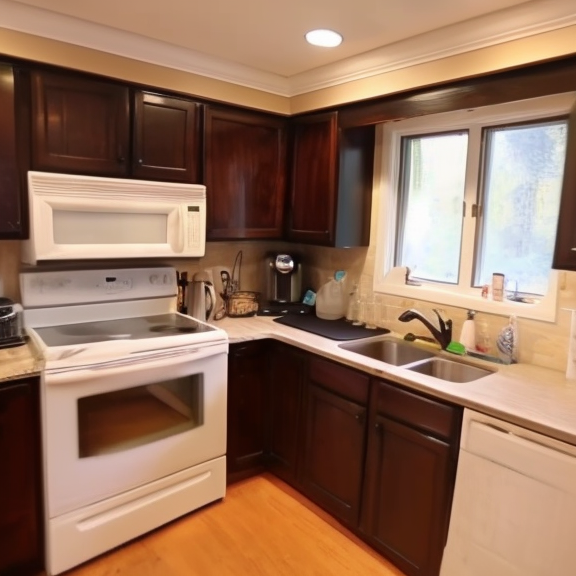} &
        \includegraphics[width=0.16\textwidth]{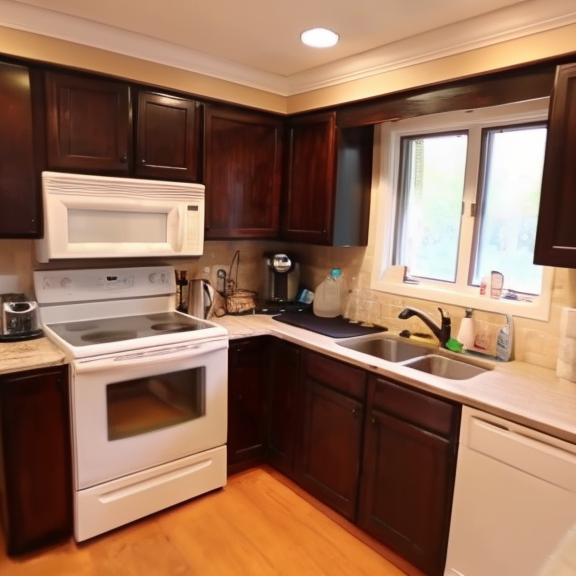} & 
        \includegraphics[width=0.16\textwidth]{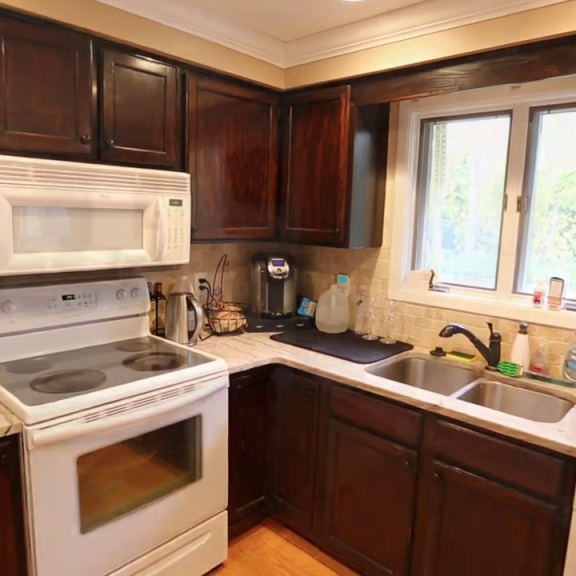} \\

        \includegraphics[width=0.16\textwidth]{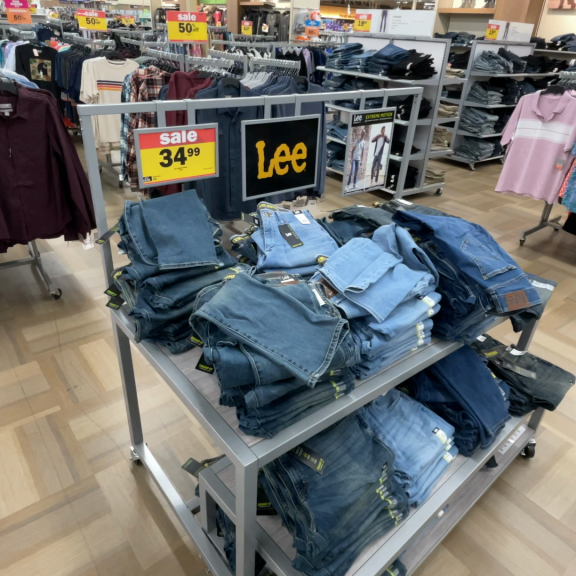} & 
        \includegraphics[width=0.16\textwidth]{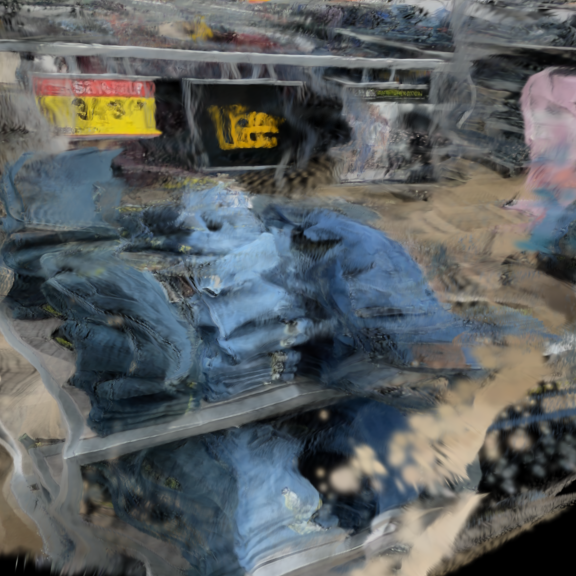} & 
        \includegraphics[width=0.16\textwidth]{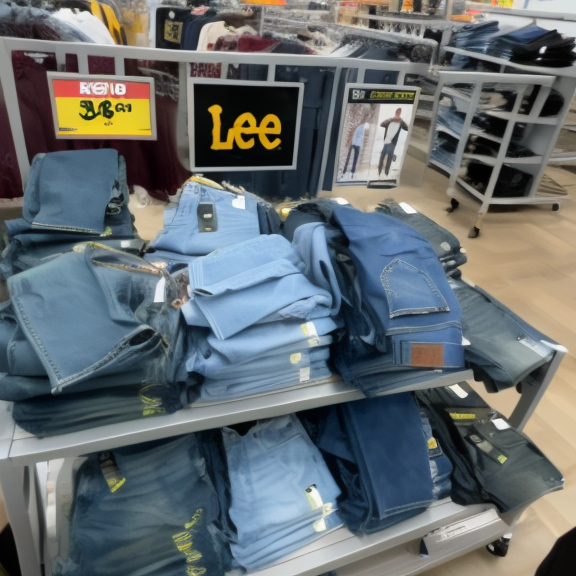} &
        \includegraphics[width=0.16\textwidth]{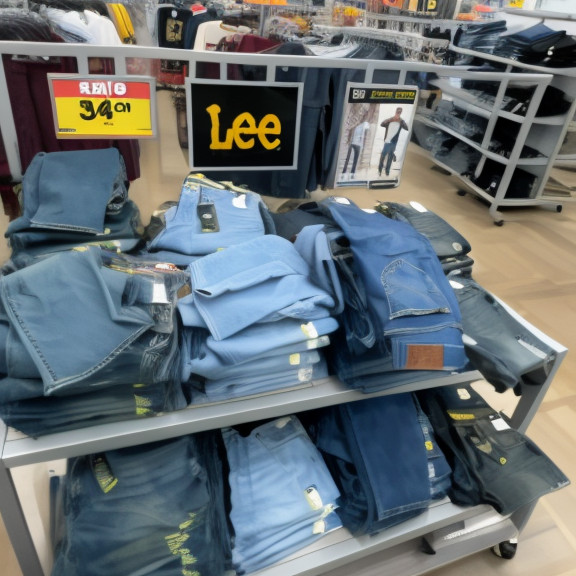} & 
        \includegraphics[width=0.16\textwidth]{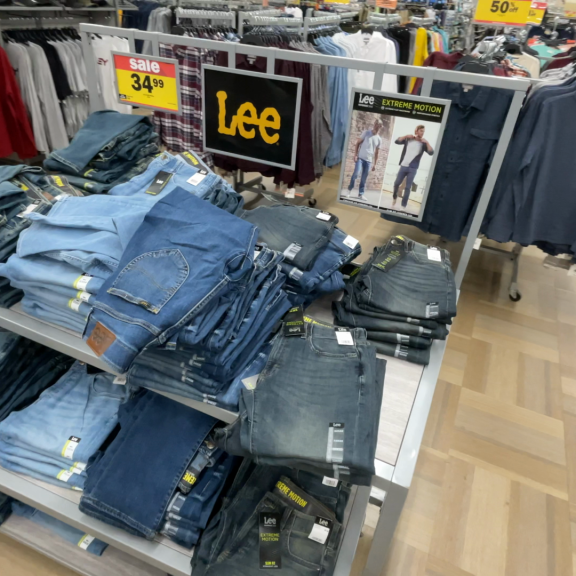} \\

        \includegraphics[width=0.16\textwidth]{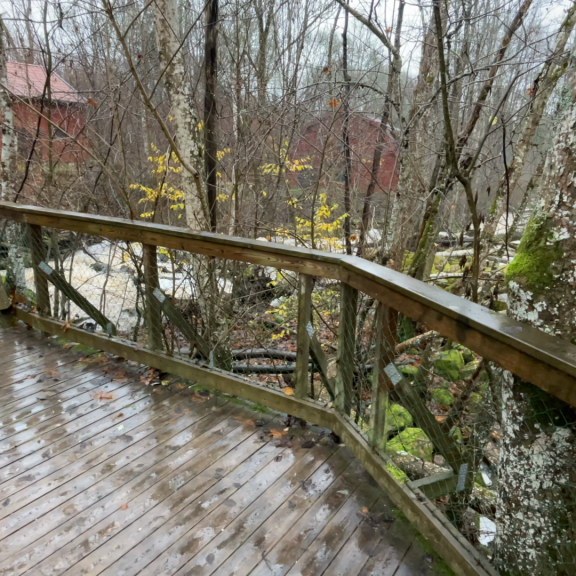} & 
        \includegraphics[width=0.16\textwidth]{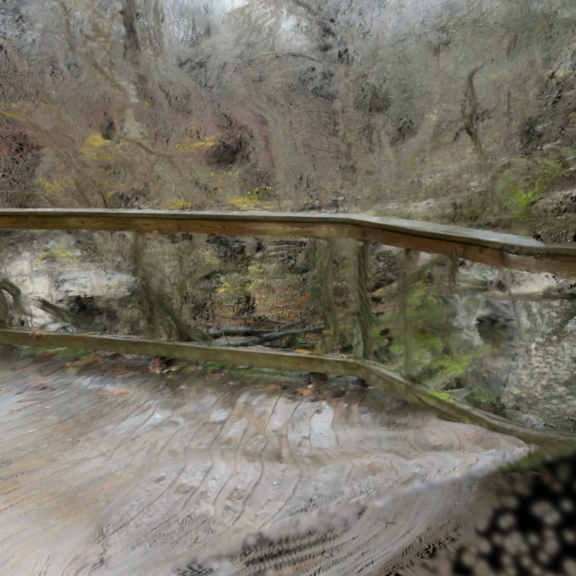} & 
        \includegraphics[width=0.16\textwidth]{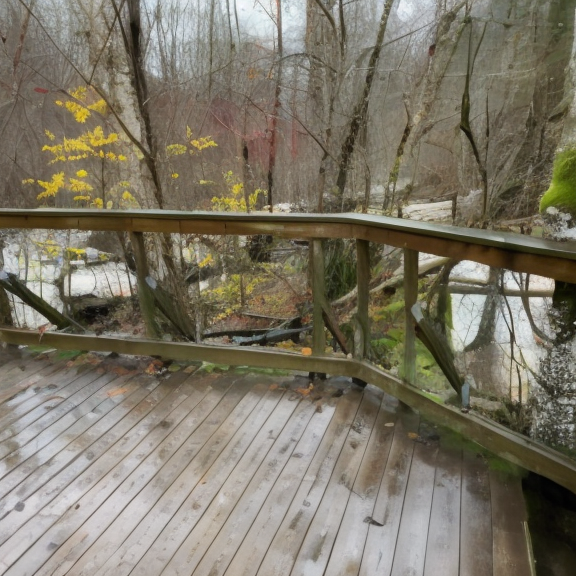} &
        \includegraphics[width=0.16\textwidth]{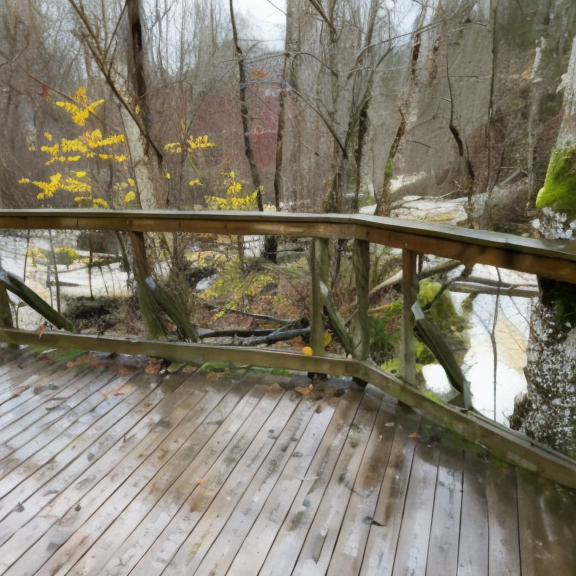} & 
        \includegraphics[width=0.16\textwidth]{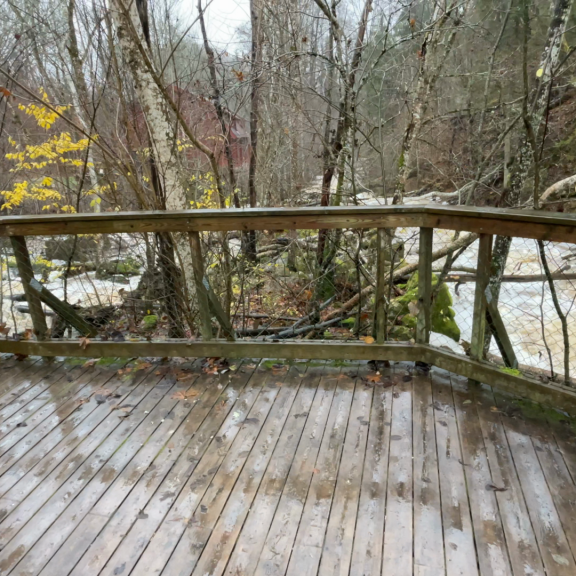} \\

        \includegraphics[width=0.16\textwidth]{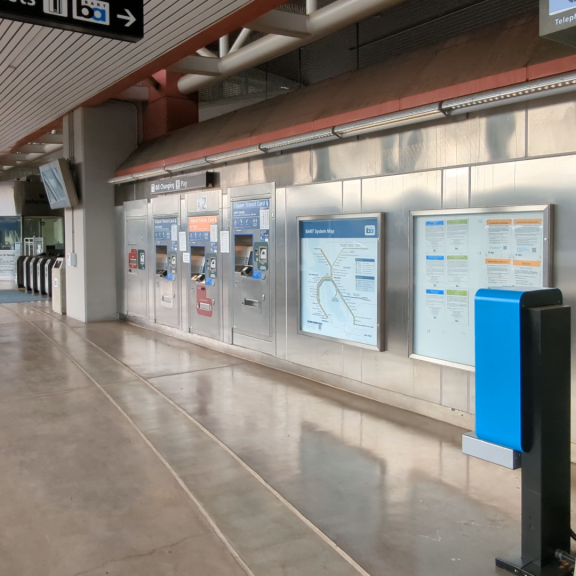} & 
        \includegraphics[width=0.16\textwidth]{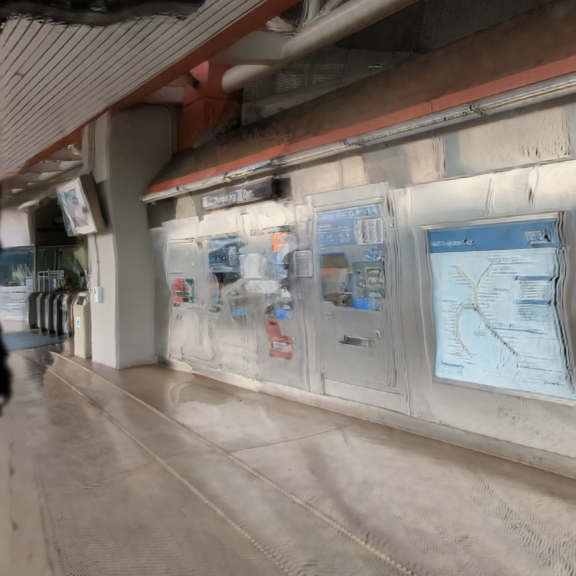} & 
        \includegraphics[width=0.16\textwidth]{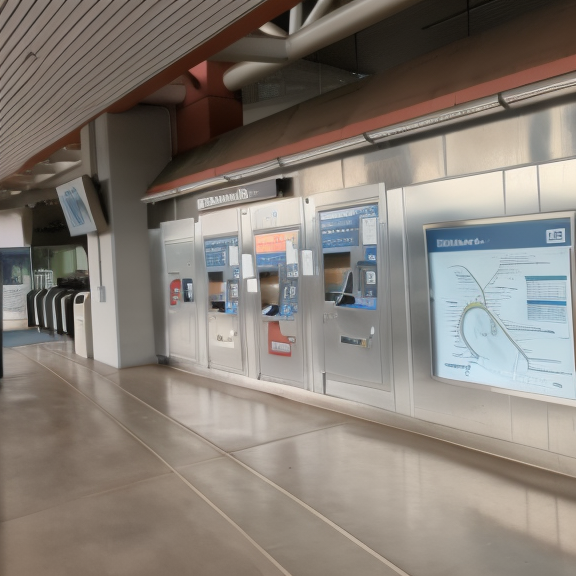} &
        \includegraphics[width=0.16\textwidth]{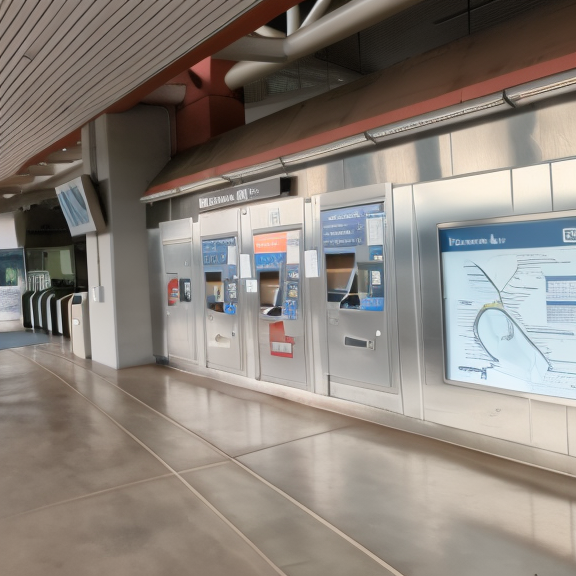} & 
        \includegraphics[width=0.16\textwidth]{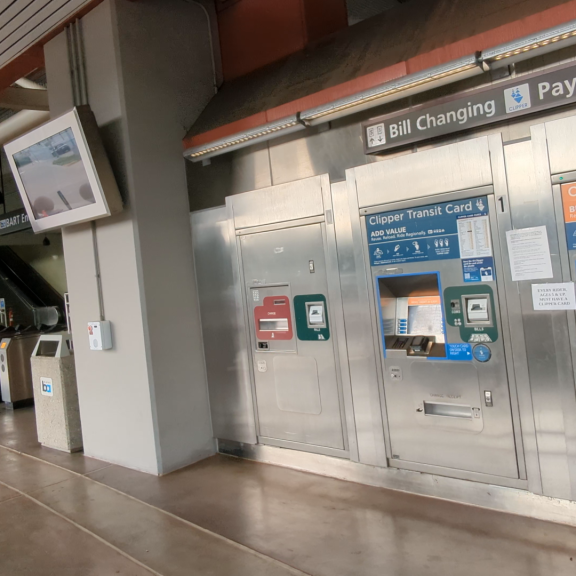} \\

        \hline
    \end{tabular}
    \vspace{-0.2cm}
    \caption{\textit{Qualitative comparison of novel view synthesis results across different methods. Each row shows target view generation between two input frames. While many frames are generated between two input frames, this figure only shows \textbf{\textit{one generated frame}} for all baselines (regression, \THISWORK, and diffusion) due to space limitation.}}
    \label{fig:appendix_images}
\end{figure*}

Figure~\ref{fig:appendix_images} presents representative examples of visual quality across different baseline methods. The input to this task consists of two camera frames (\textit{i.e.,} Input Frame 1 and Input Frame 2), and the NVS workload generates multiple intermediate frames between these two input views. For brevity, we display only one generated frame per input scene (each row represents a different scene). Compared to the diffusion baseline, which serves as our quality gold standard, the regression baseline produces inconsistent views with noticeable occlusion artifacts. In contrast, \THISWORK\ with a quality factor of 95\% consistently generates output frames with visual quality closely matching the diffusion baseline, while achieving substantially lower computational cost.

\end{document}